\documentclass[review]{elsarticle}

\usepackage{amsmath,amssymb,amsfonts}
\usepackage{algorithmic}
\usepackage{algorithm}
\usepackage{setspace}
\usepackage{subfigure}
\usepackage{multirow}
\usepackage{xcolor}
\usepackage{hyperref}

\journal{Journal of \LaTeX\ Templates}

\begin{document}

\begin{frontmatter}

\title{Automatic Preference Based Multi-objective Evolutionary Algorithm on Vehicle Fleet Maintenance Scheduling Optimization}


\author[mymainaddress]{Yali Wang}
\cortext[mycorrespondingauthor]{Corresponding author}
\ead{y.wang@liacs.leidenuniv.nl}

\author[mysecondaryaddress]{Steffen Limmer}
\ead{steffen.limmer@honda-ri.de}

\author[mysecondaryaddress]{Markus Olhofer}
\ead{Markus.Olhofer@honda-ri.de}

\author[mymainaddress]{Michael Emmerich}
\ead{m.t.m.emmerich@liacs.leidenuniv.nl}

\author[mymainaddress]{Thomas B{\"a}ck}
\ead{t.h.w.Baeck@liacs.leidenuniv.nl}

\address[mymainaddress]{Leiden Institute of Advanced Computer Science, Leiden University, 2333 CA Leiden, The Netherlands}
\address[mysecondaryaddress]{Honda Research Institute Europe GmbH, 63073 Offenbach, Germany}

\begin{abstract}
A preference based multi-objective evolutionary algorithm is proposed for generating solutions in an automatically detected knee point region. It is named Automatic Preference based DI-MOEA (AP-DI-MOEA) where DI-MOEA stands for Diversity-Indicator based Multi-Objective Evolutionary Algorithm). AP-DI-MOEA has two main characteristics: firstly, it generates the preference region automatically during the optimization; secondly, it concentrates the solution set in this preference region. Moreover, the real-world vehicle fleet maintenance scheduling optimization (VFMSO) problem is formulated, and a customized multi-objective evolutionary algorithm (MOEA) is proposed to optimize maintenance schedules of vehicle fleets based on the predicted failure distribution of the components of cars. Furthermore, the customized MOEA for VFMSO is combined with AP-DI-MOEA to find maintenance schedules in the automatically generated preference region. Experimental results on multi-objective benchmark problems and our three-objective real-world application problems show that the newly proposed algorithm can generate the preference region accurately and that it can obtain better solutions in the preference region. Especially, in many cases, under the same budget, the Pareto optimal solutions obtained by AP-DI-MOEA dominate solutions obtained by MOEAs that pursue the entire Pareto front.
\end{abstract}


\end{frontmatter}


\section{Introduction}
\label{sec:introduction}
The fleet maintenance scheduling optimization (VFMSO) problem was initially proposed in \cite{wang2019vehicle} due to the increasing demand by companies, corporations, and organizations of all sorts which rely on vehicle fleets to deliver products and services and need to maintain vehicles for safety reasons. In the problem, a vehicle fleet, such as a taxi fleet, a bus fleet, etc., can be maintained in multiple separate workshops according to a maintenance schedule. To be specific, each workshop has its own capacity and ability, meaning that on the one hand, each workshop has its own team and each team can work on only one car at the same time; on the other hand, each workshop is limited to the maintenance of the specific component(s) due to restrictions in the equipment or skill level of the staff. The maintenance schedule is optimized for each component based on its remaining useful lifetime (RUL) which has been predicted through predictive approaches or models \cite{elattar2016prognostics}. Furthermore, the cost and time which are needed by different workshops are considered because it is possible that the maintenance of the same component produces different costs and workloads when the operation is performed in different workshops. The VFMSO problem is essential because it can not only ensure the safety of vehicles for use; at the same time, it can lead to low maintenance costs and longer lives for vehicles as well.

To enhance the approach in \cite{wang2019vehicle}, to be specific, to handle the uncertainty in the problem and apply it on new application scenarios, in this paper, we improve it from the following two aspects:

\begin{itemize}
\item[1.] There exists a lot of uncertainty when we use the predicted RUL for each component as its due date, because no matter how accurate the predictive model is, it is still possible that the component will break on other dates: before the due date or later. Therefore, instead of only the RUL, we decide to involve the RUL probability distribution as the foundation to assign the maintenance time in the scheduling optimization. 

\item[2.] The VFMSO problem usually leads to a large and complex solution space, however, finding the most preferred solution is the ultimate goal. To this end, AP-DI-MOEA (Automatic Preference based DI-MOEA) is developed based on the framework of DI-MOEA (Diversity-Indicator based Multi-Objective Evolutionary Algorithm) \cite{wang2019diversity}. The new algorithm can generate the preference region automatically and find solutions with a more fine-grained resolution in the preference region.
\end{itemize}

This paper is organized as follows. Section \ref{sec:formulation} formulates the enhanced VFMSO problem. A literature review on preference based optimization is provided in Section \ref{sec:literature}. The customized multi-objective evolutionary algorithm for the enhanced VFMSO is introduced in Section \ref{sec:customizedalg}, and in Section \ref{sec:ap-di-moea}, we explain AP-DI-MOEA. Section \ref{sec:experiments} presents and discusses experiments and their results. Lastly, Section \ref{sec:conclusion} concludes the paper and outlines directions for future work.

\section{Problem Formulation}
\label{sec:formulation}
For a car fleet operated by an operator, the components of cars (e.g., springs, brakes, tires or the engine) can fail and should be maintained regularly. Some separate workshops are available for the maintenance of the car fleet, and the repair time and maintenance cost are known for each component in each workshop. Beside the time and cost for repairing the car component, a fixed set-up cost and set-up time are also considered for each visit of a car to a workshop, which correspond to the cost and time required for the preparation of the maintenance operation. 

The enhanced VFMSO problem addressed in this paper is defined as follows:
\begin{enumerate}
\item There are $n$ cars $C=\{{C_{1},C_{2}},\cdots,C_{n}\}$ and $m$ workshops $W=\{{W_{1},W_{2}},\\\cdots,W_{m}\}$.

\item Each car $C_i$ comprises $l_i$ components to be maintained for $i=1,\cdots,n$.

\item For each component $O_{ij}$ ($j=1,\cdots,l_i$), i.e., the $j$th component of car $C_i$, there is a set of workshops capable of repairing it. The set of workshops is represented by $W_{ij}$ which is a subset of $W$.

\item The processing time for maintaining component $O_{ij}$ in workshop $W_k$ is predefined and denoted by $p_{ijk}$.

\item The cost for maintaining component $O_{ij}$ in workshop $W_k$ is predefined and denoted by $q_{ijk}$.

\item The set-up time of car $C_i$ in workshop $W_k$ is predefined and denoted by $x_{ik}$.

\item The set-up cost of car $C_i$ in workshop $W_k$ is predefined and denoted by $y_{ik}$.

\item The number of teams in workshop $W_k$ is predefined and denoted by $z_k$.

\item The previous repair time of component $O_{ij}$ is recorded and denoted by $L_{ij}$.
\end{enumerate}

At the same time, the following assumptions are made:
\begin{enumerate}
\item All workshops and teams are available at the time of the optimization and assumed to be continuously available.

\item All the components are independent from each other.

\item Times required for transport of cars from/to workshops are included in the maintenance time and cost of cars, and the set-up time.

\item Environmental changes (such as car accidents) are not considered here.

\item There are no precedence constraints among the components of different cars. Cars are maintained on a first-come-first-served basis.

\item Each team can only work on one operation at a time and an operation, once started, must run to completion.

\item No operation can start before the completion of the previous operation.
\end{enumerate}

Two constraints are considered in the problem. As mentioned earlier, each workshop can only repair specific components, and this is the first constraint. Another constraint is that the maintenance periods of different operations for the same car should not overlap. It is obviously wrong if two overlapping maintenance operations of a car are assigned to different workshops because one car cannot be in two different workshops at the same time. If two overlapping maintenance operations of a car are assigned to the same workshop, it is not correct either because these two maintenance operations should be grouped together as one operation in this case. The grouping strategy will be explained in Section \ref{sec:customizedalg}.

Three objectives are assumed to be relevant for the vehicle fleet operator, which are the total workload, total cost and expected number of failures. In a multi-objective optimization problem, the objectives typically are conflicting, i.e., achieving the optimal value for one objective requires some compromise on other objectives. In our problem, the fact that a faster maintenance usually is more expensive leads to the conflict between the first two objectives. The expected number of failures counts the times when the vehicles are broken on the road. Here, the expected value is used because the actual value is unknown at the time of the optimization due to uncertainties in the predictions. When the expected number of failures is large, less maintenance tasks are performed, therefore, the workload and cost can drop. 

Let $T_k$ denote the sum of the times spent for all operations that are processed in workshop $W_k$; $M_i$ the sum of all costs spent for all maintenance operations of car $C_i$; $F_{ij}$ the number of failures of component $O_{ij}$. Three objectives can be defined as: 
\begin{flalign}
&\text{Minimize the total workload:}  ~~  f_1 = \sum_{k=1}^{m}T_k &&\\
&\text{Minimize the total cost:}    ~~ f_2 = \sum_{i=1}^{n}M_i &&\\
\nonumber
&\text{Minimize the expected number of failures:} \\
&f_3 = \sum_{i=1}^{n}\sum_{j=1}^{l_i} \mathbb{E}(F_{ij})
\end{flalign}

\section{LITERATURE REVIEW}
\label{sec:literature}

Multi-objective scheduling optimization is a major topic in the research of manufacturing systems. Its fundamental task is to organize work and workloads to achieve comprehensive optimization in multiple aspects, such as the processing time, processing cost and production safety, by deploying resources, setting maintenance time and processing sequence. In past decades, this issue has received a great deal of interest and research in different fields, such as scheduling of charging/discharging for electric vehicles \cite{zakariazadeh2014multi}; scheduling in cloud computing \cite{ramezani2015evolutionary}; scheduling of crude oil operations \cite{hou2015pareto}; scheduling in the manufacturing industry to reduce carbon emissions \cite{ding2016carbon}; scheduling medical treatments for resident patients in a hospital \cite{jeric2012multi}; scheduling for Internet service providers \cite{bhamare2017multi}, and so on.

As a typical workshop style, the flexible job shop scheduling problem (FJSP) is an essential branch of production planning problems. The FJSP consists of a set of independent jobs to be processed on multiple machines, and each job contains several operations with a predetermined order. It is assumed that each operation must be processed in specified processing time on a specific machine out of multiple alternatives. The problem has been extensively studied in the literature (for example, \cite{chiang2013simple}, \cite{yuan2015multiobjective}, \cite{gao2019review}). The FJSP is the research basis of the maintenance scheduling optimization problem and many real-world problems extend the standard FJSP by adding specific features. \cite{ozguven2010mathematical} considers FJSP-PPF (process plan flexibility), where jobs can have alternative process plans. It is assumed that the process plans are known in advance and that they are represented by linear precedence relationships. Because only one of the alternative plans has to be adopted for each job, the FJSP-PPF deals with not only routing and sequencing sub-problems, but also the process plan selection sub-problem. In this paper, a mixed-integer linear programming model is developed for the FJSP-PPF. In \cite{demir2014effective}, a mathematical model and a genetic algorithm are proposed to handle the feature of overlapping in operations. It is assumed that a lot which contains a batch of identical items is transferred from one machine to the next only when all items in the lot have completed their processing, therefore, sublots are transferred from one machine to the next for processing without waiting for the entire lot to be processed at the predecessor machine, meaning that starting a successor operation of job is not necessary to finish of its predecessor completely. Three features are considered in \cite{yu2017extended}, which are (1) job priority; (2) parallel operations: some operations can be processed simultaneously; (3) sequence flexibility: the sequence of some operations can be exchanged. A mixed integer liner programming formulation (MILP) model is established to formulate the problem and an improved differential evolution algorithm is designed. Because of unexpected events occurring in most of the real manufacturing systems, there is a new type of scheduling problem known as the dynamic scheduling problem. This type of problem considers random machine breakdowns, adding new machine, new job arrival, job cancellation, changing processing time, rush order, rework or quality problem, due date changing, etc. Corresponding works on the FJSP include  \cite{fattahi2010dynamic}, \cite{al2011robust}, \cite{shen2015mathematical}, \cite{ahmadi2016multi}. Compared with the standard FJSP, our VFMSO problem has some special properties: (1) flexible sequence: the sequence of the components is not predefined, but mainly influenced by the RUL probability distribution. (2) multiple problem parameters: besides the processing time, other problem parameters like the maintenance cost, set-up time, set-up cost, repair teams, etc, also have impacts on the result.


Our real-world problem, like many other multi-objective optimization problems, can lead to a large objective space. However, finding a well-distributed set of solutions on the Pareto front requests a large population size and computational effort. Therefore, instead of spreading a limited size of individuals across the entire Pareto front, we decide to only focus on a part of the Pareto front, to be specific, the search for solutions will be only guided towards the preference region which, in our algorithm, is determined by the knee point. It has been argued in the literature that knee points are most interesting solutions, naturally preferred solutions and most likely the optimal choice of the decision maker (DM) \cite{das1999characterizing, mattson2002minimal, deb2003multi, branke2004finding}.  



The knee point is a point for which a small improvement in any objective would lead to a large deterioration in at least one other objective. In the last decade, several methods have been presented to identify knee points or knee regions. Das \cite{das1999characterizing} refers the point where the Pareto surface ``bulges" the most as the knee point, and this point corresponds to the farthest solution from the convex hull of individual minima which is the minima of the single objective functions. Zitzler \cite{zitzler2004tutorial} defines $\epsilon$-dominance: a solution $a$ is said to $\epsilon$-dominate a solution $b$ if and only if $f_i(a)+\epsilon \geq f_i(b) ~\forall i=1,...,m$ where $m$ is the number of objectives. A solution with a higher $\epsilon$-dominance value with respect to the other solutions in the Pareto front approximation, is a solution having higher trade-offs and in this definition corresponds to a knee point. The authors of \cite{yu2018method} propose to calculate the density of solutions projected onto the hyperplane constructed by the extreme points of the non-dominated solutions, then identify the knee regions based on the solution density. 

Different algorithms of applying knee points in MOEA have also been proposed.
Branke \cite{branke2004finding} modifies the second criterion in NSGA-II \cite{deb2002fast}, and replaces the crowding distance by either an angle-based measure or a utility-based measure. The angle-based method calculates the angle between an individual and its two neighbors in the objective space. The smaller the angle, the more clearly the individual can be classified as a knee point. However, this method can only be used for two objective problems. In the utility-based method, a marginal utility function is suggested to approximate the angle-based measure in the case of more than two objectives. The larger the external angle between a solution and its neighbors, the larger the gain in terms of linear utility obtained from substituting the neighbors with the solution of interest. However, the utility-based measure is not suited for finding knees in concave regions of the Pareto front.

Rachmawati \cite{rachmawati2006multi, rachmawati2006multi2} proposes a knee-based MOEA which computes a transformation of original objective values based on a weighted sum niching approach. The extent and the density of coverage of the knee regions are controllable by the parameters for the niche strength and pool size. The strategy is susceptible to the loss of less pronounced knee regions.

Sch{\"u}tze \cite{schutze2008approximating} investigates two strategies for the approximation of knees of bi-objective optimization problems with stochastic search algorithms. Several new definitions for identifying knee points and knee regions for bi-objective optimization problems has been suggested in \cite{deb2011understanding} and the possibility of applying them has also been discussed.

Besides the knee points, the reference points, which are normally provided by the DM, have also been used to find a set of solutions near reference points. Deb \cite{deb2006reference} proposes an MOEA, called R-NSGA-II, by which a set of Pareto optimal solutions near a supplied set of reference points can be found. The dominance relation together with a modified crowding distance operator is used in this methodology. For all solutions of the population, the distances to all reference points are calculated and ranked. The lowest rank (over all reference points) of a solution is used as its crowding distance. Besides, a parameter $\epsilon$ is used to control the spread of obtained solutions. Bechikh proposes KR-NSGA-II \cite{bechikh2010searching} by extending R-NSGA-II. Instead of obtaining the reference points from the DM, in KR-NSGA-II, the knee points are used as mobile reference points and the search of the algorithm was guided towards these points. The number of knee points of the optimization problem is needed as prior information in KR-NSGA-II.

Gaudrie \cite{gaudrie2019targeting} uses the projection (intersection in case of a continuous front) of the closest non-dominated point on the line connecting the estimated ideal and nadir points as default preference. Conditional Gaussian process simulations are performed to create possible Pareto fronts, each of which defines a sample for the ideal and the nadir point, and the estimated ideal and nadir are the medians of the samples.

Rachmawati and Srinivasan \cite{rachmawati2009multiobjective} evaluate the worthiness of each non-dominated solution in terms of compromise between the objectives. The local maxima is then identified as potential knee solutions and the linear weighted-sums of the original objective functions are optimized to guide solutions toward the knee regions. 

Another idea of incorporating preference information into evolutionary multi-objective optimization is proposed in \cite{thiele2009preference}. They combine the fitness function and an achievement scalarizing function containing the reference point. In this approach, the preference information is given in the form of a reference point and an indicator-based evolutionary algorithm IBEA \cite{zitzler2004indicator} is modified by embedding the preference information into the indicator. Various further preference based MOEAs have been suggested, e.g., \cite{braun2011preference, ramirez2017knee, wang2017new}. 

In our proposed algorithm, i.e., AP-DI-MOEA, we adopt the method from \cite{das1999characterizing} to identify the knee point, design the preference region based on the knee point, and guide the search towards the preference region. The advantages of our algorithm are: (1) no prior knowledge is used in identifying the knee point and knee region; (2) the preference region is generated automatically and narrowed down step by step to benefit its accuracy; (3) our strategy cannot only handle bi-objective optimization problems, but also tri- and many-objective problems; (4) although we integrate the strategy with DI-MOEA, it may be integrated with any standard MOEAs (such as NSGA-II \cite{deb2002fast}, SMS-EMOA \cite{beume2007sms} and others); (5) the proposed algorithm is capable of finding preferred solutions for multi-objective optimization problems with linear, convex, concave Pareto fronts and discrete problems.

\section{Customized Algorithm for Vehicle Fleet Maintenance Scheduling Optimization}
\label{sec:customizedalg}
For our real-world VFMSO problem, we first define the execution window for each component based on its predicted RUL probability distribution which is assumed to be a normal distribution. The execution window suggests that the maintenance of the component can only start at a time spot inside the window. The mean ($\mu$) and standard deviation ($\sigma$) of the predicted RUL probability distribution determine the interval of the execution window, which is defined as: [$\mu -2\times \sigma$, $\mu +2\times \sigma$]. The interval is chosen relatively long because 95\% of the values are within two standard deviations of the mean, therefore, maintenance before or after the interval hardly makes sense.

After the determination of the execution window, the following two special strategies have been taken to improve the process of scheduling optimization: 
\begin{itemize}
\item Grouping components.
\item Obtaining the penalty cost and expected number of failures by Monte Carlo simulation.
\end{itemize}
Lastly, evolutionary algorithm (EA) is chosen to solve this real-world application problem due to its powerful characteristics of robustness and flexibility to capture global solutions of complex combinatorial optimization problems. Moreover, EAs are well suited to solve multi-objective optimization problems due to their ability to approximate the entire Pareto front in a single run.

\subsection{Grouping Components}
It would be troublesome and also a waste of time and effort to send a car to workshops repeatedly in a short period of time to repair different components. In our algorithm, since each component has its execution window for its maintenance, it is possible to combine the maintenance of several components to one visit if their execution windows overlap. Especially, by grouping the maintenance of multiple components into one maintenance operation, the set-up cost and set-up time are charged only once for the complete group of components. 

\begin{figure}[!htbp]
\centering
\includegraphics[height=120pt, width=260pt]{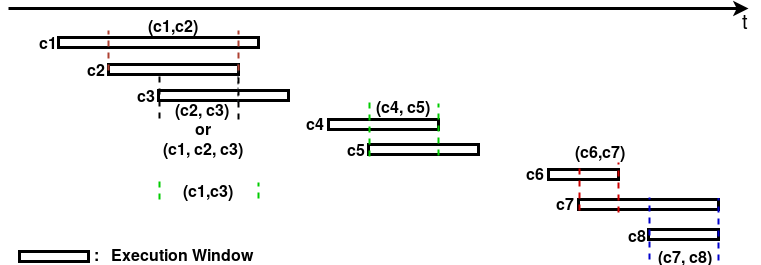} 
\caption{Possible groups for a car with eight components.} 
\label{f2}
\end{figure}

Figure~\ref{f2} represents the execution windows of eight components of a car. The overlap of the execution windows shows the possibility of grouping these components. Combining components can only be effective if there is a common overlap of the execution window for components from the same car, and the starting time of the group operation must lie within the common overlap. In this example, component $c1$ can be grouped with $c2$ and/or $c3$ due to the overlap between their execution windows. Other possible group structures can be deduced in the same manner. 

\subsection{Monte Carlo Simulation}
\begin{figure}[!htbp]
\centering
\includegraphics[ width=4in]{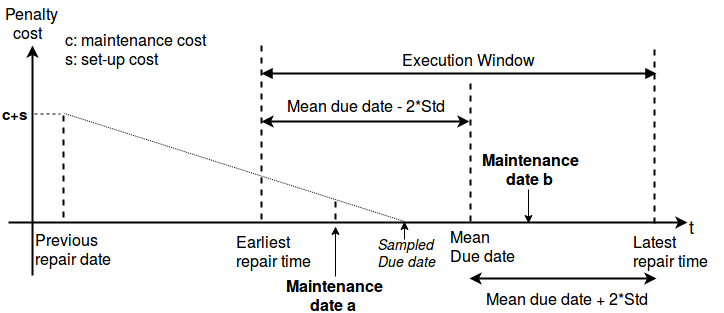} 
\caption{Execution window of a component.} 
\label{f1}
\end{figure}

Within the execution window of a component, an arbitrary time can be chosen as the starting time for maintaining the component. The maintenance time of each component should be as close as possible to its real due date, because:
\begin{itemize}
\item Performing the maintenance too early results in higher maintenance costs in the long term, because more maintenance tasks have to be done. 
\item The risk of breaking down on the road will increase if the maintenance date is too late.
\end{itemize}
Therefore, we use Monte Carlo simulation to simulate the ``real'' due dates for each component. In our experiments, stability can be achieved at a few hundred samples, in our case, $1000$ samples of the due date are generated in the execution window of each component according to its predicted RUL probability distribution (see Section IV). Figure~\ref{f1} shows an example of the execution window evolved from the predicted RUL probability distribution of a component. After 1000 sampled due dates are generated in the execution window, the scheduled maintenance date of the component is compared with these samples one by one, and each comparison can lead to three situations. Let us use $d_{ij}^v$  to denote the $v$th due date sample of component $O_{ij}$; and $D_{ij}$ the scheduled maintenance date of component $O_{ij}$. Three possibilities after the comparison are:
\\
Case 1) $~D_{ij} < d_{ij}^v$\\
The scheduled maintenance date is earlier than the sample (or the ``real'' due date) means that the component will be maintained before it is broken. In this case, its useful life between the maintenance date and the due date will be wasted. Therefore, a corresponding penalty cost is imposed to reflect the waste. To calculate the penalty cost, a linear penalty function is suggested based on the following assumptions:
\begin{itemize}
\item If a component is maintained when it is new or the previous maintenance has just completed, the penalty cost would be the full cost of maintaining it, which is $c+s$: the maintenance cost of the component and the set-up cost of the car;
\item If a component is maintained at exactly its due date, the penalty cost would be 0.
\end{itemize}

\vspace{-0.0cm}

Assume $d_{ij}^v$ is ``Sampled Due date'' in Figure~\ref{f1}, and $D_{ij}$ is ``Maintenance date a'', in this case, $D_{ij}$ is earlier than $d_{ij}^v$. The penalty cost of ``Maintenance date a'' for ``Sampled Due date'' would be the vertical dotted line above ``Maintenance date a''.
\\
Case 2) $~D_{ij} > d_{ij}^v$\\
The scheduled maintenance date is later than the sample means that the maintenance date is too late and the defect occurs on the use. Still, $d_{ij}^v$ is ``Sampled Due date'' in Figure~\ref{f1}, but the scheduled maintenance date $D_{ij}$ is ``Maintenance date b''. In this case, $D_{ij}$ is later than $d_{ij}^v$, and the vehicle will break down on the road. In our algorithm, the number of failures will be increased by one.
\\
Case 3) $~D_{ij} = d_{ij}^v$\\
The ideal situation is that the maintenance date is scheduled on the due date. The component can be maintained exactly at the date that the component is broken. In this case, there is no penalty or failure.

The averages of the penalty costs and the number of failures from $1000$ due date samples will be used as the penalty cost and expected number of failures for the scheduled maintenance date of the component. For each operation (the single-component operation or group operation), its cost consists of three parts: the set-up cost of the car, the maintenance costs and the penalty costs of all components of the operation. The penalty cost of components is a part of the total cost, and the expected number of failures of components is the third objective to be minimized in our multi-objective optimization.

\subsection{Implementation of Evolutionary Algorithm Operators}
To solve our application problem with an EA, there are several basic issues we need to deal with, such as, how to represent an individual or solution in the population (Chromosome Encoding); how to take these chromosomes into a process of evolution (Genotype-Phenotype Mapping); how to create variations of solutions in each iteration (Genetic Operators). Details of these topics are given in the following subsections.

\subsubsection{Chromosome Encoding}
\begin{figure*}[!htbp]
\hspace{-0.7cm}
\includegraphics[height=0.45in, width=5.2in]{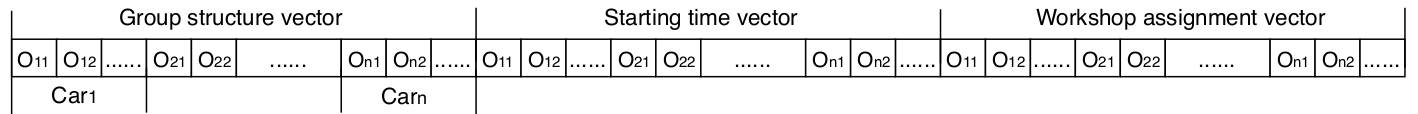}
\caption{Three-vector chromosome.}
\label{f0}
\end{figure*}

In our algorithm, a three-vector chromosome (Figure~\ref{f0}) is proposed to represent an individual, and the three vectors are:
\begin{itemize}
  \item Group structure vector: the group structures of components.
  \item Starting time vector: the starting times of operations.
  \item Workshop assignment vector: the workshops for operations.
\end{itemize}

The group structure vector gives the information of which components are in the same group, it is initialized by randomly picking a feasible group structure for each car (check the details in \cite{wang2019vehicle}). The generation of the starting time vector should be later than the generation of the group structure vector because the starting time of each operation is determined by the execution window which is the entire execution window of the component for a single-component operation or the execution window intersection for a group operation. A time spot is randomly selected from the execution window or execution window intersection for each operation in order to initialize the starting time vector. 

A workshop is considered as ``several workshops" based on its capacity (the number of teams). By this way, the schedule for each workshop team can be achieved from the solution. For example, consider that two workshops have three and four repairing teams respectively. Then, group operations can be randomly assigned to seven ``workshops'', the former three and the latter four represent the corresponding teams in two workshops.

\subsubsection{Genotype-Phenotype Mapping}
To use the power of EAs to obtain a better population, we need to evaluate each chromosome and give the better ones higher probabilities to produce offspring. This is done by genotype-phenotype mapping or decoding the chromosome. In our problem, it is to convert an individual into a feasible schedule to calculate the objectives and constraints which represent the relative superiority of a chromosome. The genotype-phenotype mapping can be easily achieved in our algorithm because the group structure, the starting time and the workshop team of the operations can be acquired directly from each individual. When converting an individual into a schedule, it is possible that the processing times of two or more operations assigned to the same workshop team are overlapping since the starting time of each operation is decided in the starting time vector. In this situation, the principle of first-come-first-served is followed: the starting time and processing time of the earlier started operation remain the same; the starting time of the later started operation is delayed until the completion of the previous operation; the processing time of the later started operation remains the same; while, an extra waiting time is added to the later started operation as a penalty because the vehicle waits in the workshop for the maintenance.

\subsubsection{Genetic Operators}
In accordance with the problem and its encoding, specific crossover and mutation operators have been designed for our problem (check the details in \cite{wang2019vehicle}). Both operators are applied separately to the three parts of the chromosome.

For the group structure vector, multi-point crossover can be used as crossover operator and the number of cutting points depends on the length of the vector. The same cutting points can be applied to the starting time vector when performing crossover. However, the change on the group structure vector as a consequence of the crossover may result in the invalidity of genes in the starting time vector because it is possible that the group members and execution window intersections have changed due to the new group structure. Therefore, when performing the crossover on the starting time vector, the starting times of all operations should be checked based on the new group structure and a new starting time is produced randomly from the correct intersection in the case that the starting time of an operation is invalid. The multi-point crossover can be applied to the workshop assignment vector as well. 

The mutation operator alters one or more gene values in a chromosome. Similarly, the mutation should be operated on the group structure vector first due to its impact on the starting time vector; the starting time of operations should be checked and corrected after the mutation is done on the group structure vector. Afterwards, several gene values can be altered in the staring time vector and workshop assignment vector to generate a new individual.

\section{Proposed Preference based Algorithm}
\label{sec:ap-di-moea}
As the number of objectives and decision variables increases, the number of non-dominated solutions tends to grow exponentially \cite{pal2018decor}. This brings more challenges on achieving efficiently a solution set with satisfactory convergence and diversity. At the same time, a huge number of solutions is needed to approximate the entire Pareto front. However, a big population means more computational time and resources. To overcome these difficulties, we propose an automatic preference based MOEA, which can generate the preference region or the region of interest (ROI) automatically and find non-dominated solutions in the preference region instead of the entire Pareto front. The automatic preference based MOEA is developed based on the framework of DI-MOEA (Diversity-Indicator based Multi-Objective Evolutionary Algorithm) \cite{wang2019diversity}. We call our new algorithm AP-DI-MOEA.
 
DI-MOEA is an indicator-based MOEA, it has shown to be competitive to other MOEAs on common multi-objective benchmark problems. Moreover, it is invariant to the shape of the Pareto front and can achieve evenly spread Pareto front approximations.
DI-MOEA adopts a hybrid selection scheme:
\begin{itemize}
 \item The ($\mu$ + $\mu$) generational selection operator is used when the parent population can be layered into multiple dominance ranks. The intention is to accelerate convergence until all solutions are non-dominated.
 \item The ($\mu$ + 1) steady state selection operator is adopted in the case that all solutions in the parent population are mutually non-dominated and the diversity is the main selection criterion to achieve a uniform distribution of the solutions on the Pareto front.
\end{itemize}

DI-MOEA employs non-dominated sorting as the first ranking criterion; the diversity indicator, i.e., the Euclidean distance based geometric mean gap indicator, as the second, diversity-based ranking criterion to guide the search. Two variants of DI-MOEA, denoted as DI-1 and DI-2, exist, which use the crowding distance and diversity indicator, respectively, as the second criteria in the ($\mu$ + $\mu$) generational selection operator. While, to ensure the uniformity of the final solution set, the diversity indicator is used by both variants in the ($\mu$ + 1) steady state selection operator. Analogously, two variants of AP-DI-MOEA, i.e., AP-DI-1 and AP-DI-2, are derived from the two variants of DI-MOEA.

The workings of AP-DI-MOEA are outlined in Algorithm 1. (Exceedance of) $Enum\_P$ is a predefined condition (In our algorithm, $Enum\_P$ is the number of evaluations.) to divide the algorithm into two phases: learning phase and decision phase. In the learning phase, the algorithm explores the possible area of Pareto optimal solutions and finds the rough approximations of the Pareto front. In the decision phase, the algorithm identifies the preference region and finds preferred solutions. When the algorithm starts running and satisfies $Enum\_P$ at some moment, the first preference region will be generated and  $Enum\_P$ will be updated for determining a new future moment when the preference region needs to be updated. The process of updating $Enum\_P$ continues until the end. The first $Enum\_P$ is a boundary line. Before it is satisfied, AP-DI-MOEA runs exactly like DI-MOEA to approximate the whole Pareto front; while, after it is satisfied, the preference region is generated automatically and AP-DI-MOEA finds solutions focusing on the preference region. The subsequent values of $Enum\_P$ define the later moments to update the preference region step by step, eventually, a precise ROI with a proper size can be achieved.

\begin{figure*}[!htbp]
\vspace{-3.5cm}
\hspace{-3cm}
\includegraphics[height=10in]{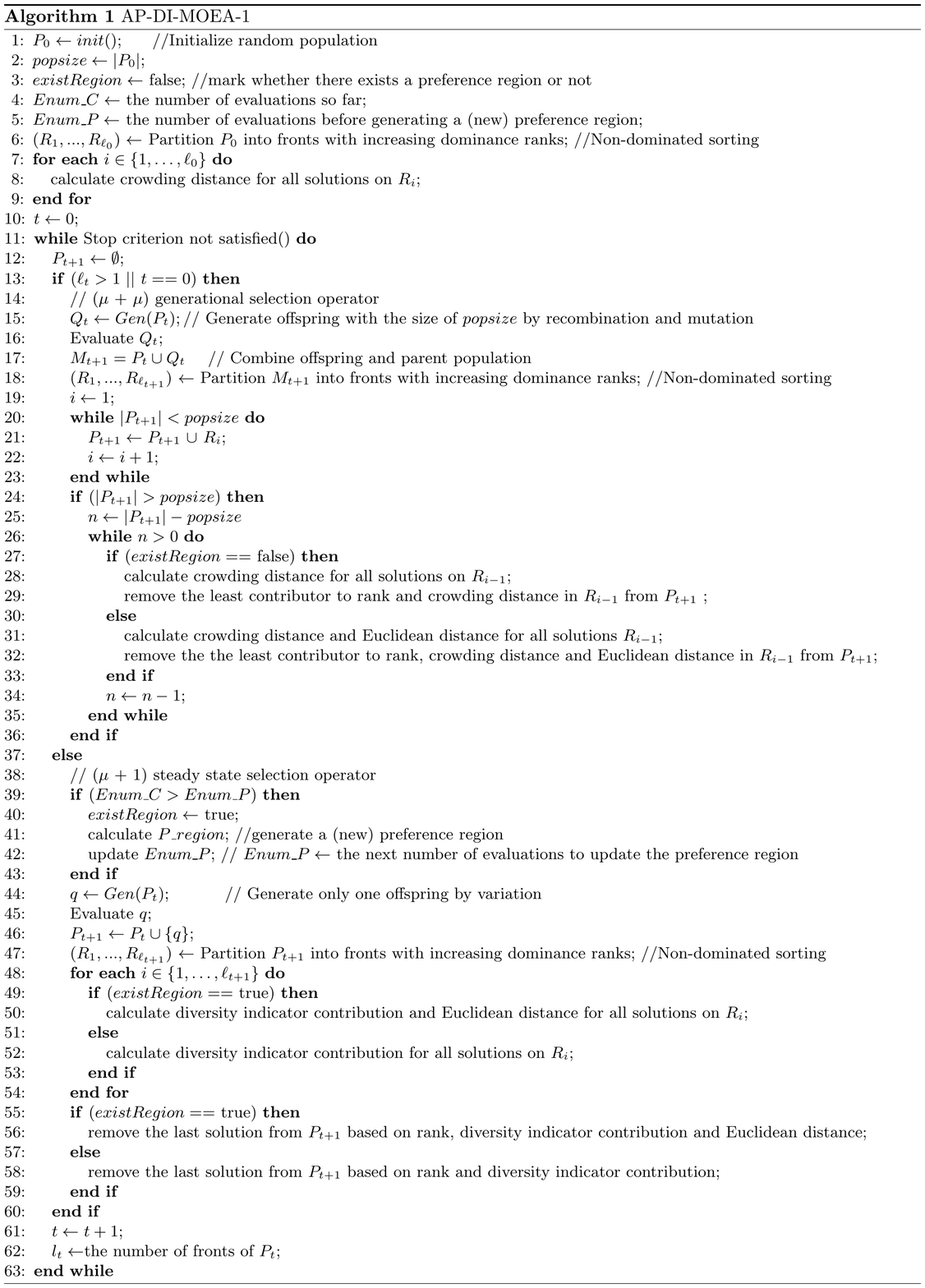}
\end{figure*}

\begin{figure*}[!htbp]
\hspace{-2.5cm}
\includegraphics[width=7in]{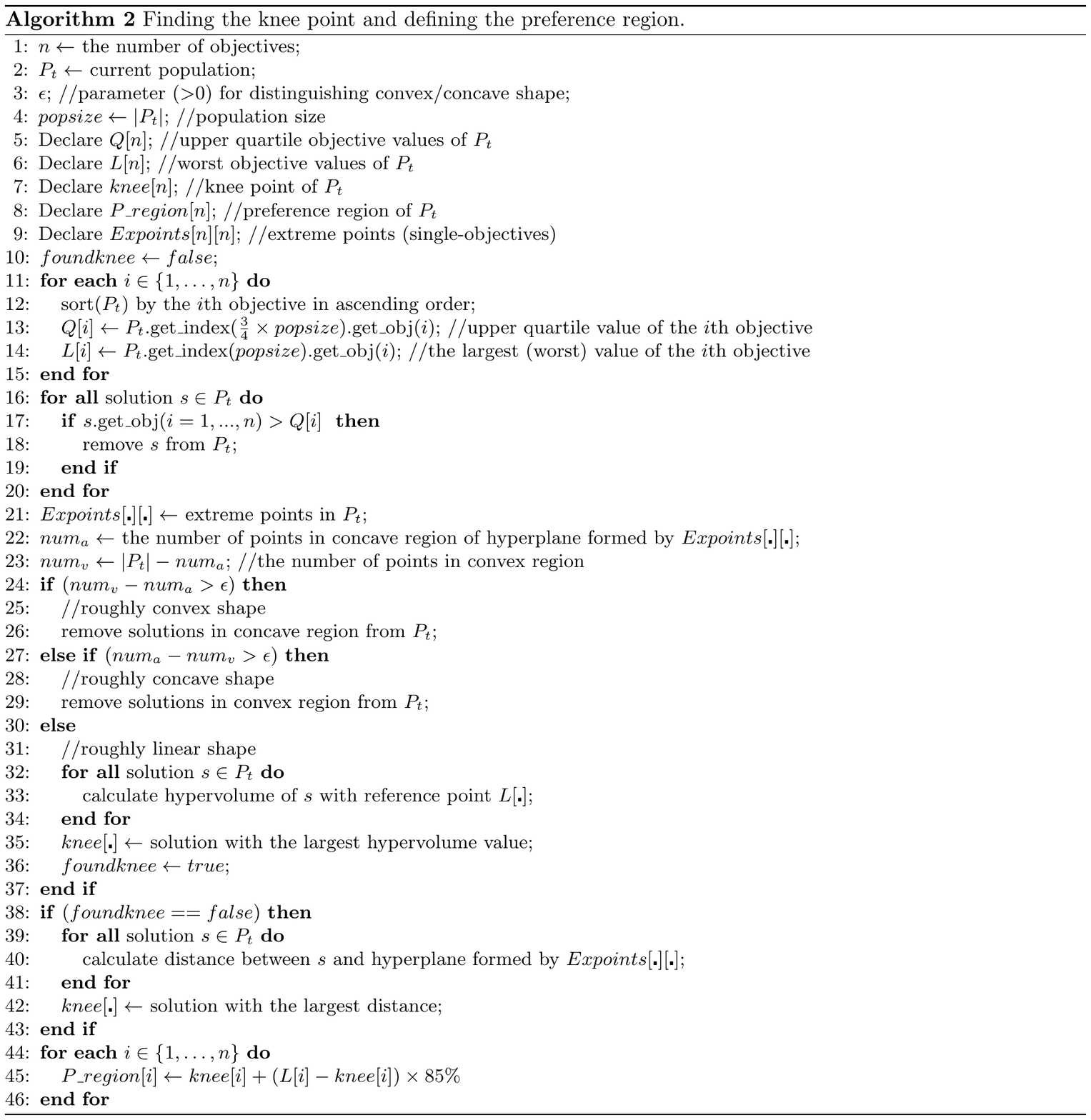}
\end{figure*}

\begin{figure}[!htbp]
\centering
\includegraphics[width=3.3in]{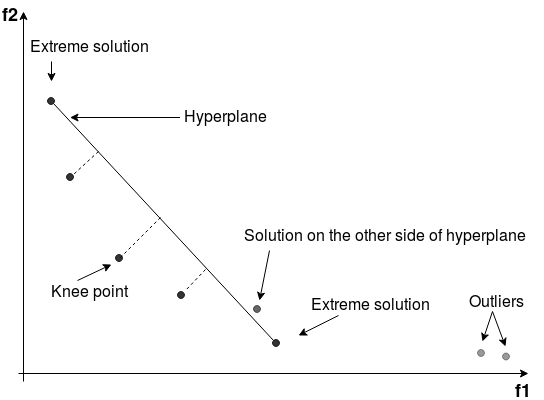} 
\caption{Finding the knee point in bi-dimensional space.} 
\label{knee}
\end{figure}

The first/new preference region is formed based on the population at the moment when the condition of $Enum\_P$ is satisfied, especially the knee point of the population. Algorithm 2 gives the details of line 41 in Algorithm 1, it introduces the steps of finding the knee point of a non-dominated solution set and constituting a hypercube shaped preference region according to the knee point. Figure~\ref{knee} also gives an illustration of finding the knee point in bi-dimensional space. Firstly, the upper quartile objective values (line 13 in Algorithm 2) in the solution set are used as a boundary to define outliers and solutions outside this boundary are removed  (line 16-20 in Algorithm 2). The extreme solutions (the solutions with the maximum value in one objective) (line 21 in Algorithm 2) are then found inside the boundary and a hyperplane is formed based on the extreme solutions. In a bi-dimensional space (Figure~\ref{knee}), the hyperplane is only a line connecting two extreme solutions. According to the numbers of points below and above the hyperplane (line 22 - 23 in Algorithm 2), the shape of the solution set can be roughly perceived.
We will distinguish between ``convex'' and ``concave'' regions. Points in the \textit{convex} (\textit{concave}) \textit{region} are dominating (dominated by) at least one point in the hyperplane spanned by the extreme points. However, when the number of the points in the convex region and the number of points in the concave region is close enough, it implies that the shape of the current solution set is almost linear. This occurs both when the true Pareto front is linear and when the solution set is converged very well in a small area of the Pareto front. A parameter $\epsilon$ then is used to represent the closeness and it is a small number decided by the size of the solution set. In the case that the shape of the current solution set is (almost) linear, the solution with the largest hypervolume value with regards to the worst objective vector (line 14 in Algorithm 2) is adopted as the knee point (line 32 - 36 in Algorithm 2). While, under the condition that the shape of the current solution set is convex or concave, the solution in the convex or concave region with the largest Euclidean distance to the hyperplane is chosen as the knee point (line 39 - 42 in Algorithm 2). After the knee point is found, the preference region can be determined based on the knee point by the following formula:
\begin{align}
P\_region[i] = knee[i] + (L[i]-knee[i]) \times 85\%
\end{align}

Let $i$ denotes the $i$th objective, as in Algorithm 2, $L[i]$ is the worst value of the $i$th objective in the population, $knee[i]$ is the $i$th objective value of the knee point and $P\_region[i]$ is the upper bound of the $i$th objective. W.l.o.g. we assume the objectives are to be minimized and the lower bound of preference region is the origin point. According to the formula, we can see that the first preference region is relatively large (roughly 85\% of the entire Pareto front). With the increase in the number of iteration, the preference region will be updated and becomes smaller and smaller because every preference region picks 85\% of the current Pareto front. Eventually, we want the preference region can reach a proper range, say, 15\% of the initial Pareto front. The process of narrowing down the preference region step by step can benefit the accuracy of the preference region.

In the interest of clarity, Algorithm 1 only shows the workings of AP-DI-1, the workings of AP-DI-2 can be obtained by replacing crowding distance with the diversity indicator contribution.
In the ($\mu$ + $\mu$) generational selection operator (line 14 - 36 in Algorithm 1), when there is no preference region, the second ranking criteria (the crowding distance for AP-DI-1; the diversity indicator for AP-DI-2) for all solutions on the last front are calculated and the population will be truncated based on non dominated sorting and the second ranking criteria (line 28 - 29 in Algorithm 1). While, if a preference region already exists, both the second ranking criteria and Euclidean distance to the knee point for all solutions on the last front are calculated and the population will be truncated based on first non dominated sorting, then the second ranking criteria, lastly, Euclidean distance to the knee point (line 31 - 32 in Algorithm 1). In the ($\mu$ + 1) steady state selection operator (line 38 - 59 in Algorithm 1), firstly, the value of $Enum\_P$ is compared with the current number of evaluations to determine if a (new) preference region should be generated. When it is time to do so, the preference region is generated through Algorithm 2 (line 41 in Algorithm 1), at the same time, the value of $Enum\_P$ is updated to the next moment when the preference region is to be updated (line 42 in Algorithm 1). There are different strategies to assign the values of $Enum\_P$. In our algorithm, we divide the whole computing budget into two parts, the first half is used to find an initial entire Pareto front approximation, and the second half is used to update the preference region and find solutions in the preference region. Assume the total computing budget is $Enum\_T$ (the number of evaluations), then the first value of $Enum\_P$ is $\frac{1}{2}\times Enum\_T$. Due to the reason that we expect a final preference region with a size of around 15\% of the initial entire Pareto front and each new preference region takes 85\% of the current Pareto front, according to the formula: $0.85^{12} \approx 0.14$, the value of $Enum\_P$ can be updated by the following formula:
\begin{align}
Enum\_P = Enum\_P + (Enum\_T/2)/12
\end{align}

Another half of budget can be divided into $12$ partial-budgets and a new preference region is constituted after each partial-budget. In the end, the final preference region is achieved and solutions focusing on this preference region are obtained. For the rest part of the ($\mu$ + 1) steady state selection operator, likewise, when there is a preference region, three ranking criteria (1. non-dominated sorting; 2. diversity indicator; 3. the Euclidean distance to the knee point) work together to achieve a well-converged and well-distributed set of Pareto optimal solutions in the preference region.

\section{Experimental Results}
\label{sec:experiments}
\subsection{Experimental Design}
In this section, simulations are conducted to demonstrate the performance of proposed algorithms on both benchmark problems and our real-world application problems. All experiments are implemented based on the MOEA Framework (\url{http://www.moeaframework.org/}), which is a Java-based framework for multi-objective optimization.

For the two variants of AP-DI-MOEA: AP-DI-1 and AP-DI-2, their performances have been compared with DI-MOEA: DI-1, DI-2 and NSGA-III \cite{deb2014evolutionary}. We compare our algorithm with NSGA-III because NSGA-III is a representative state-of-the-art evolutionary multi-objective algorithm and it is very powerful to handle problems with non-linear characteristics. For bi-objective benchmark problems, algorithms are tested on ZDT1 and ZDT2 with 30 variables. For three objective benchmark problems, DTLZ1 with 7 variables and DTLZ2 with 12 variables are tested. For the real-world application problem of VFMSO, experiments have been conducted on two instances with different sizes. The configurations of the two instances, such as the predicted RUL probability distribution, the processing time and maintenance cost of each component, the set-up time and cost of each car, are made available on \url{http://moda.liacs.nl}. On every problem, we run each algorithm $30$ times with different seeds, while the same $30$ different seeds are used for all algorithms. All the experiments are performed with a population size of $100$; and for bi-objective problems, experiments are run with a budget of $22000$ (objective function) evaluations, DTLZ three objective problems with a budget of $120000$ evaluations, the VFMSO problems with a budget of $1200000$ evaluations. This setting is chosen to be more realistic in the light of the applications in scheduling that we ultimately want to solve.


\subsection{Experiments on bi-objective problems}

\begin{figure}[htbp]
\hspace{-0.58cm}
\subfigure[ZDT1]{
    \begin{minipage}[t]{0.53\linewidth}
        \centering
        \includegraphics[width=2in]{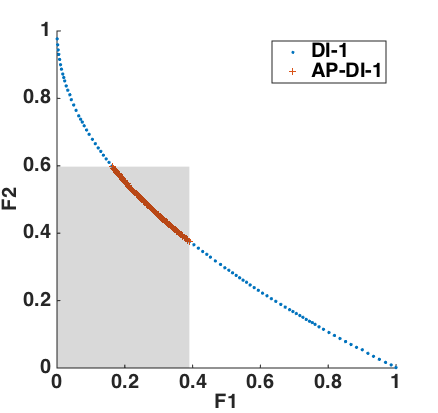}\\
        \vspace{0.02cm}
        \includegraphics[width=2in]{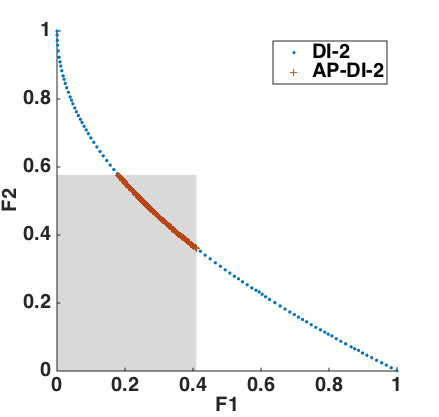}\\
        \vspace{0.02cm}
    \end{minipage}%
}%
\subfigure[ZDT2]{
    \begin{minipage}[t]{0.53\linewidth}
        \centering
        \includegraphics[width=2in]{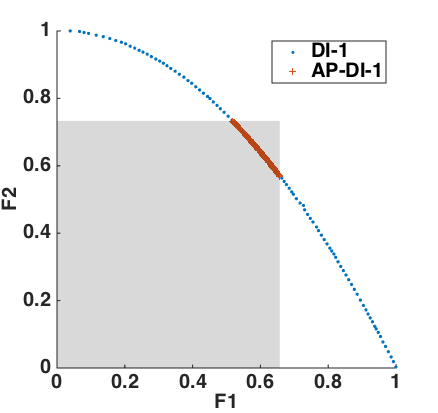}\\
        \vspace{0.02cm}
        \includegraphics[width=2in]{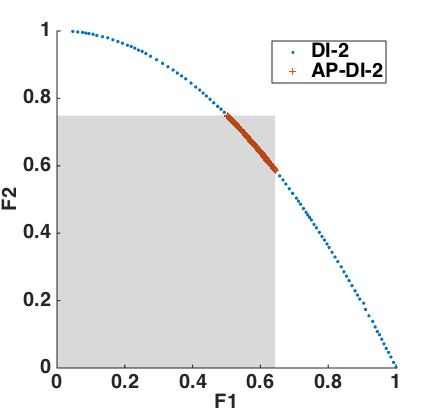}\\
        \vspace{0.02cm}
    \end{minipage}%
}%

\caption{Pareto front approximation on ZDT1 and ZDT2.}
\label{fig:ZDT1}
\end{figure}

Bi-objective problems are optimized with a total budget of $22000$ evaluations, when the number of evaluations reaches $10000$ times, the first preference region is generated, then after every $1200$ evaluations, the preference region will be updated. Figure~\ref{fig:ZDT1} shows the Pareto front approximations from a typical run on ZDT1 (left column) and ZDT2 (right column). The graphs on the upper row are obtained from DI-1 and AP-DI-1, while the graphs on the lower row are from DI-2 and AP-DI-2. In each graph, the entire Pareto front approximation from DI-MOEA and the preferred solutions from AP-DI-MOEA (or \textit{AP solutions}) are presented, at the same time, the preference region of AP-DI-MOEA is also shown by the gray area.

Besides the visualization of the Pareto fronts, we also compute the knee point of the entire final Pareto front approximation from DI-MOEA via the strategy described in Algorithm 2. For each run of DI-MOEA and AP-DI-MOEA with the same seed, the following two issues have been checked: 
\begin{itemize}
\item If the knee point from DI-MOEA is in the preference region achieved by its derived AP-DI-MOEA;
\item If the knee point from DI-MOEA is dominated by or dominating AP solutions; or if it is a non-dominated solution (mutually non-dominated with all AP solutions).
\end{itemize}

Table~\ref{table-kneezdt1} shows the results of 30 runs. For ZDT1 problem, all 30 knee points from DI-1 and DI-2 are in the preference regions from AP-DI-1 and AP-DI-2 respectively; in all these knee points, 10 from DI-1 and 7 from DI-2 are dominated by AP solutions. For ZDT2 problem, most knee points are not in corresponding preference regions, but for those in the preference regions, almost all of them are dominated by AP solutions. Please note that when a knee point from DI-MOEA is outside of the preference region from AP-DI-MOEA, it is not possible that it can dominate any AP solutions because all AP solutions are in the preference region and only solutions in the left side of the gray area can dominate AP solutions. 


\begin{table}[!htbp]
\centering
\caption{Space and dominance relation of knee point from DI-MOEA and AP solutions on ZDT problems.}
\begin{tabular}{|c|c|c|c|c|c|}
\hline
\multicolumn{2}{|c|}{Problem} & \multicolumn{2}{|c|}{ZDT1}  & \multicolumn{2}{|c|}{ZDT2}\\ 
\cline{1-6}
\multicolumn{2}{|c|}{\multirow{2}*{Algorithm}}& DI-1/ & DI-2/ & DI-1/ & DI-2/ \\
\multicolumn{2}{|c|}{ }& AP-DI-1 & AP-DI-2 & AP-DI-1 & AP-DI-2 \\
\hline
In & Incomparable & 20 & 23 & 1 & 1\\
\cline{2-6}
preference & Dominated & 10 & 7 & 9 & 9 \\
 \cline{2-6}
region & Dominating & 0 & 0 & 0 &  0 \\
\hline
Outside & Incomparable &  0 & 0 & 20 & 20 \\
\cline{2-6}
p-region & Dominated &  0 & 0 & 0 & 0 \\
\hline
\end{tabular}
\label{table-kneezdt1}
\end{table} 

We also perform the same comparison between AP-DI-MOEA and NSGA-III, the results are shown in Table~\ref{table-kneezdt1-nsga3}. For ZDT1 problem, all knee points from NSGA-III are in the preference regions from AP-DI-MOEA. Some of these knee points dominate AP solutions. For ZDT2 problem, most knee points from NSGA-III are not in the preference regions and these knee points are incomparable with AP solutions. For the knee points in the preference regions, all three dominating relations with AP solutions appear. For both problems, when the knee point from NSGA-III is dominating AP solutions, it only dominates one AP solution.

\begin{table}[!htbp]
\centering
\caption{Space and dominance relation of knee point from NSGA-III and AP solutions on ZDT problems.}
\begin{tabular}{|c|c|c|c|c|c|}
\hline
\multicolumn{2}{|c|}{Problem} & \multicolumn{2}{|c|}{ZDT1}  & \multicolumn{2}{|c|}{ZDT2}\\ 
\cline{1-6}
\multicolumn{2}{|c|}{\multirow{2}*{Algorithm}}& NSGA-III/ & NSGA-III/ & NSGA-III/ & NSGA-III/ \\
\multicolumn{2}{|c|}{ }& AP-DI-1 & AP-DI-2 & AP-DI-1 & AP-DI-2 \\
\hline
In & Incomparable & 14 & 19 & 3 & 1\\
\cline{2-6}
preference & Dominated & 0 & 0 & 2 & 3 \\
 \cline{2-6}
region & Dominating & 16 & 11 & 4 &  6 \\
\hline
Outside & Incomparable &  0 & 0 & 21 & 20 \\
\cline{2-6}
p-region & Dominated &  0 & 0 & 0 & 0 \\
\hline
\end{tabular}
\label{table-kneezdt1-nsga3}
\end{table} 

\begin{figure}[htbp]
\hspace{-0.58cm}
\subfigure[ZDT1]{
    \begin{minipage}[t]{0.53\linewidth}
        \centering
        \includegraphics[width=2in]{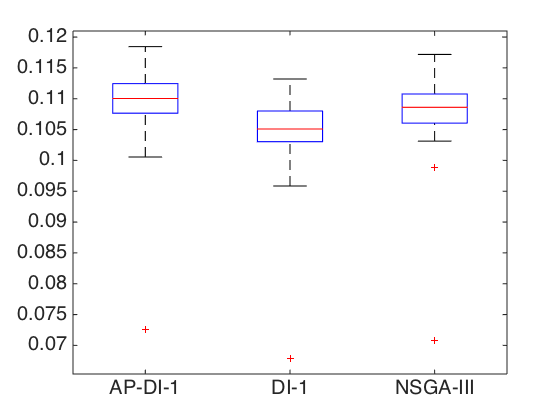}\\
        \vspace{0.02cm}
        \includegraphics[width=2in]{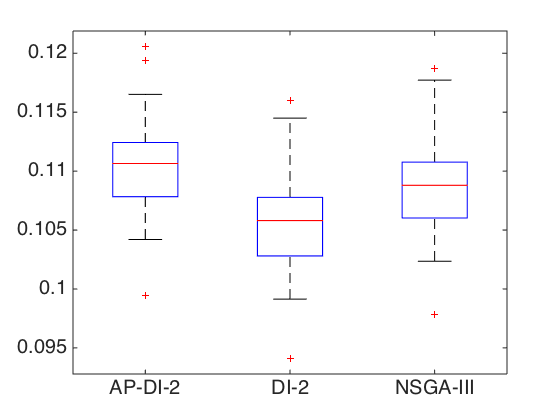}\\
        \vspace{0.02cm}
    \end{minipage}%
}%
\subfigure[ZDT2]{
    \begin{minipage}[t]{0.53\linewidth}
        \centering
        \includegraphics[width=2in]{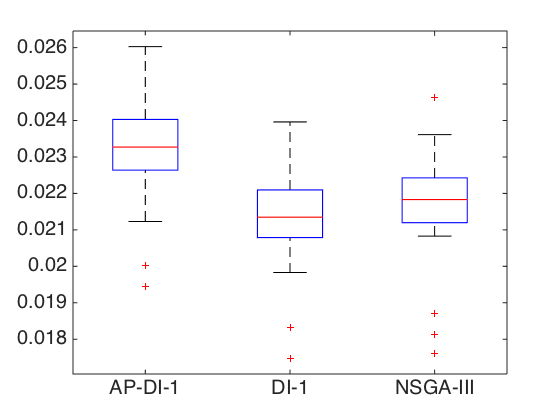}\\
        \vspace{0.02cm}
        \includegraphics[width=2in]{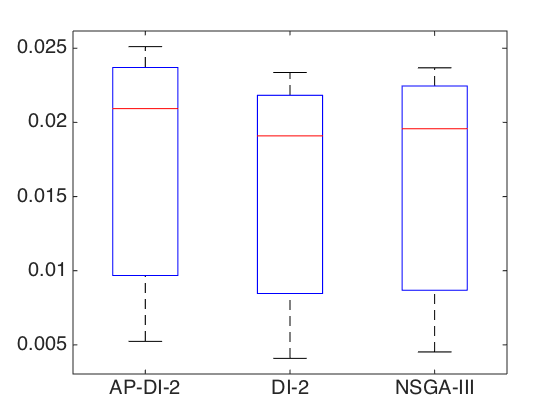}\\
        \vspace{0.02cm}
    \end{minipage}%
}%
\caption{Boxplots comparing the hypervolume values on ZDT1 and ZDT2.}
\label{box:ZDT}
\end{figure}

Instead of spreading the population across the entire Pareto front, we only focus on the preference region. To ensure that our algorithm can guide the search towards the preference region and the achieved solution set is distributed across the preference region, we compare the performance of AP-DI-MOEA, DI-MOEA and NSGA-III in the preference region. For each Pareto front approximation from DI-MOEA and NSGA-III, the solutions in the corresponding preference region from AP-DI-MOEA are picked, and we compare these solutions with AP solutions through the hypervolume indicator. The point formed by the largest objective values over all solutions in the preference region is adopted as the reference point when calculating the hypervolume indicator. It has been found that all hypervolume values of new solution sets from DI-MOEA and NSGA-III in the preference region are worse than the hypervolume values of the solution sets from AP-DI-MOEA, which proves that the mechanism indeed works in practice. Figure~\ref{box:ZDT} shows box plots of the distribution of hypervolume indicators over 30 runs.

\subsection{Experiments on three objective problems}
\begin{figure*}[htbp]
\hspace{-3.5cm}
\subfigure[DTLZ1 3 objective problem]{
    \begin{minipage}[t]{0.5\linewidth}
        \centering
        \includegraphics[height=2.5in,width=4.4in]{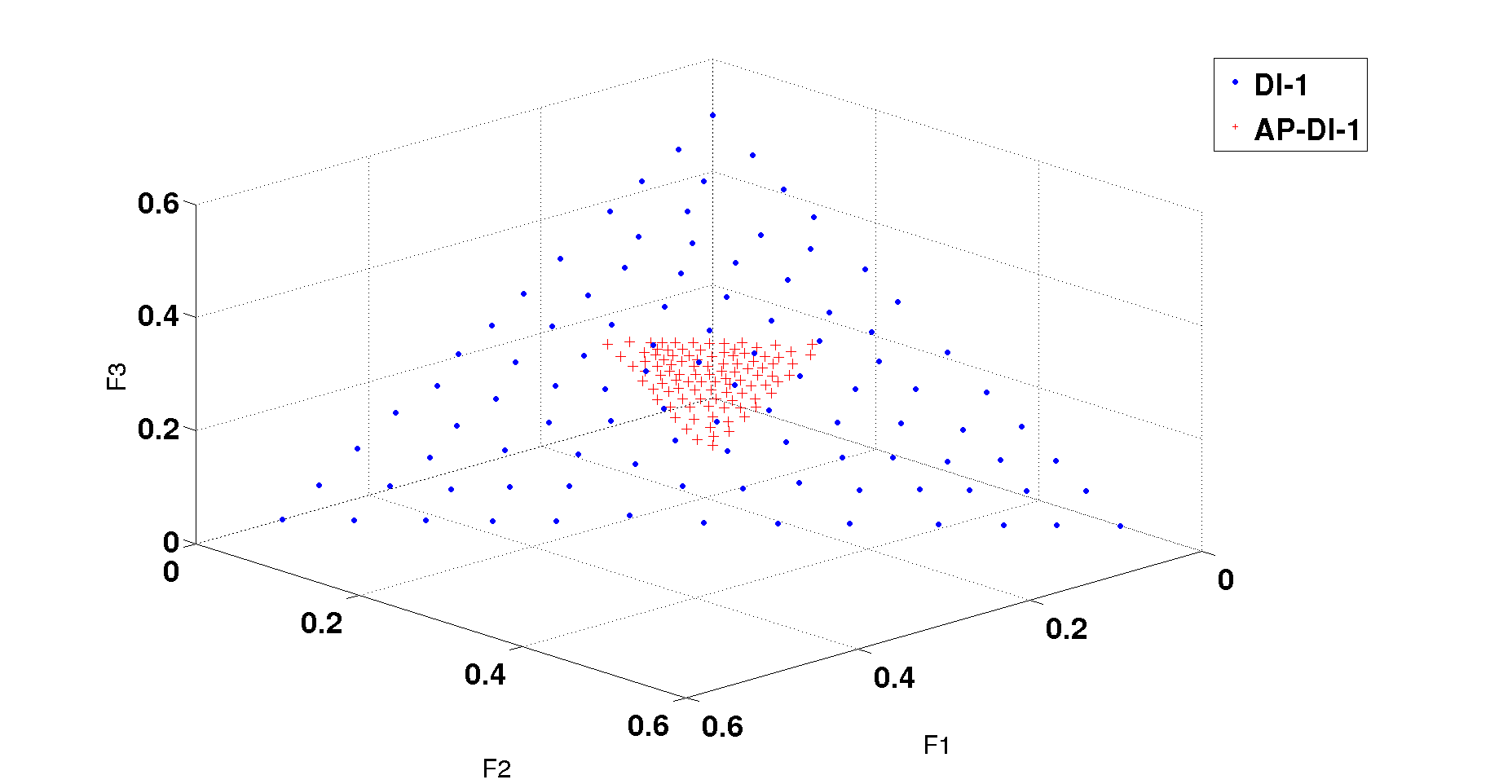}\\
        \vspace{0.45cm}
        \includegraphics[height=2.5in,width=4.4in]{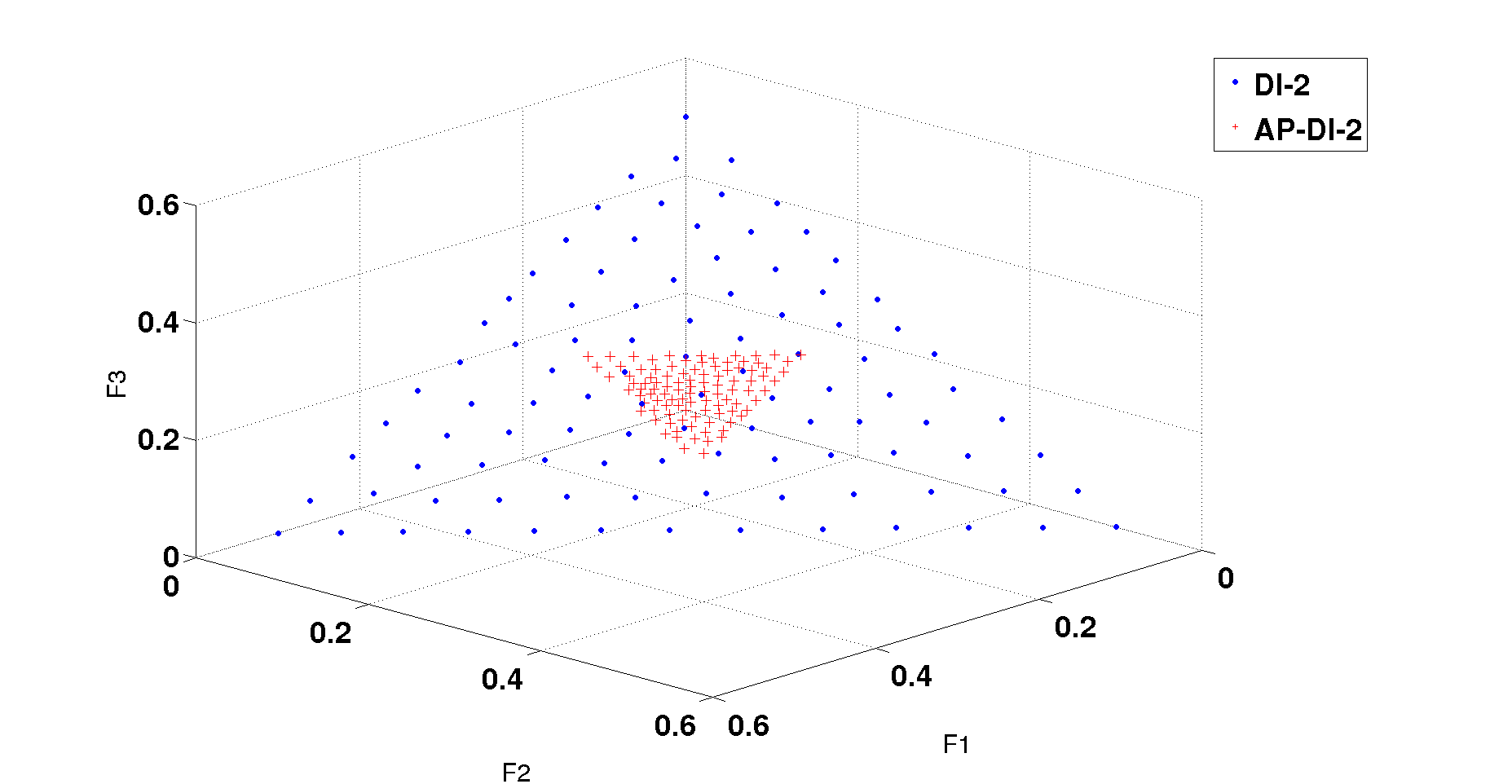}\\
    \end{minipage}%
}%
\hspace{2cm}
\subfigure[DTLZ2 3 objective problem]{
    \begin{minipage}[t]{0.5\linewidth}
        \centering
        \includegraphics[height=2.7in,width=4.4in]{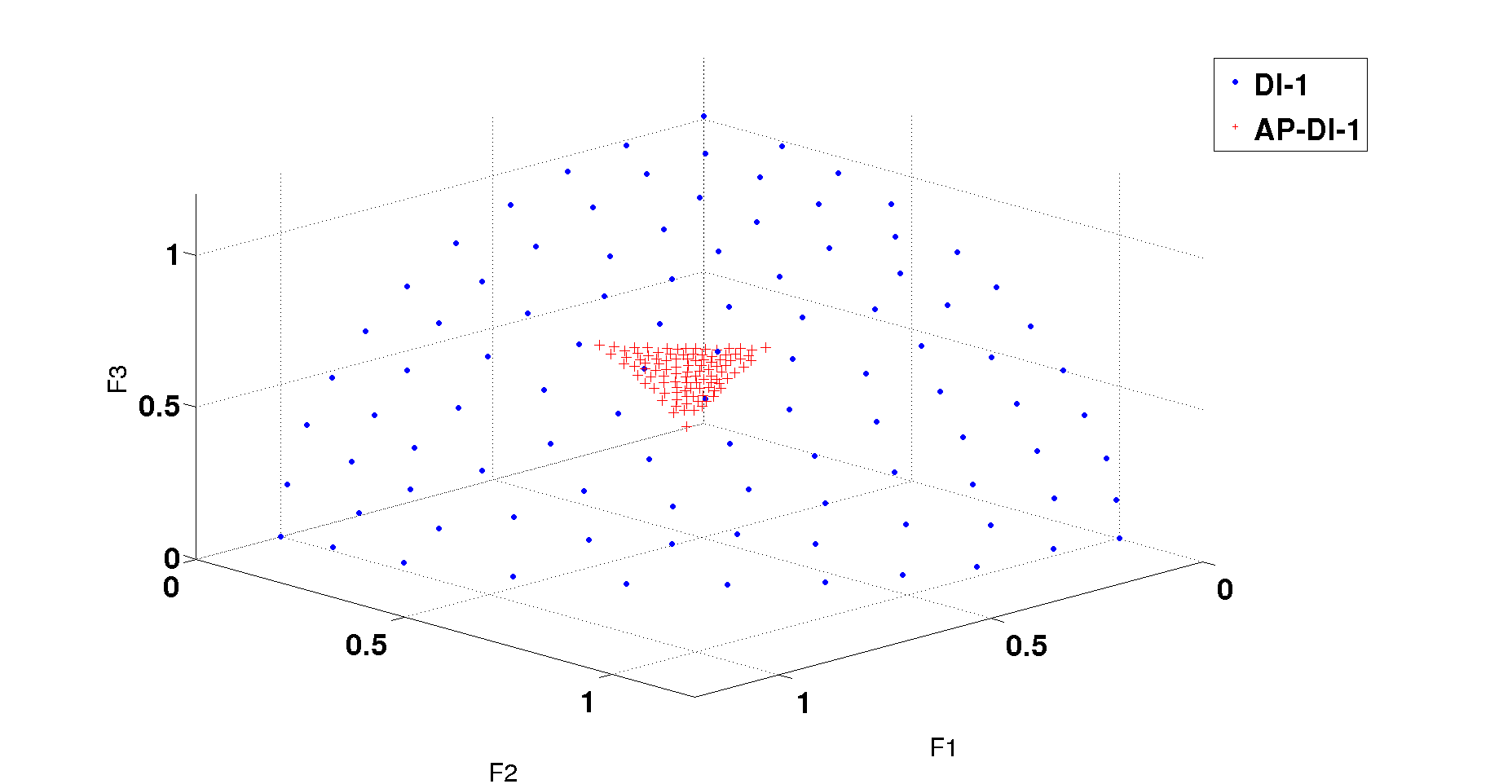}\\
        \includegraphics[height=2.7in,width=4.4in]{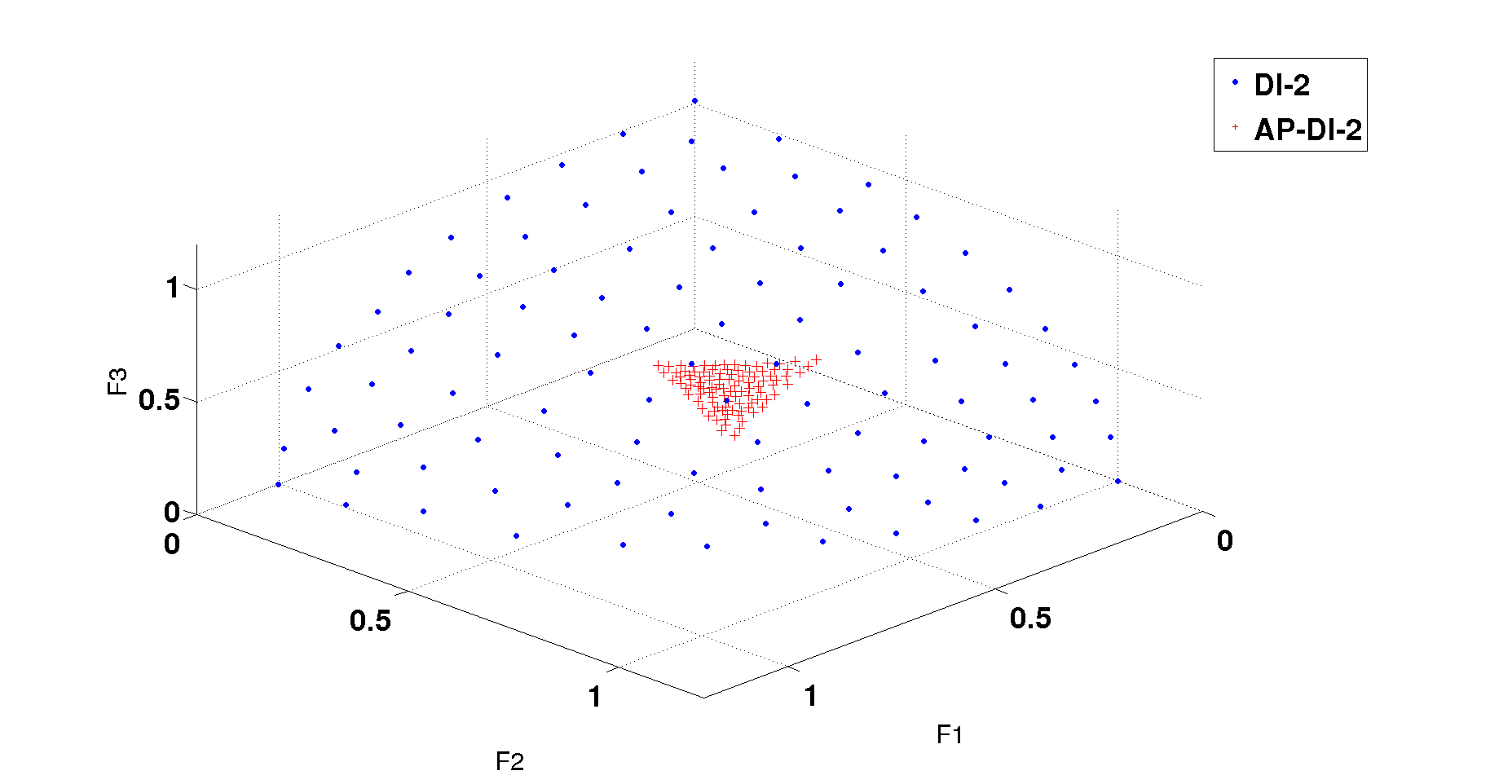}\\
    \end{minipage}%
}%
\caption{Pareto front approximation on DTLZ1 and DTLZ2.}
\label{fig:dtlz}
\end{figure*}

DTLZ1 and DTLZ2 are chosen as three objective benchmark problems to investigate our algorithms. They are performed with a total budget of $120000$ fitness evaluations, when the evaluation reaches $60000$ times, the first preference region is formed, then after every $5000$ evaluations, the preference region is updated. Figure~\ref{fig:dtlz} shows the Pareto front approximations from a typical run on DTLZ1 (left column) and DTLZ2 (right column). The upper graphs are obtained from DI-1 and AP-DI-1, while the lower graphs are from DI-2 and AP-DI-2. In each graph, the Pareto front approximations from DI-MOEA and corresponding AP-DI-MOEA are given. Since the target region is actually an axis aligned box, the obtained knee region (i.e., the intersection of the axis aligned box with the Pareto front) has an inverted triangle shape for these two benchmark problems.

Table~\ref{table-kneedtlz} shows the space and dominance relation of the knee point from DI-MOEA and the solution set from AP-DI-MOEA over 30 runs. For DTLZ1 problem, most knee points from DI-MOEA are in their respective preference regions and all knee points are mutually non-dominated with AP solutions. For DTLZ2 problem, we observed that more knee points are not in the corresponding preference regions. This is because too few solutions from DI-MOEA are in the preference region. For DTLZ1 problem, six solutions from DI-MOEA are in the corresponding preference region on average for each run, while, for DTLZ2 problem, only less than two solutions are in the corresponding preference region on average. Therefore, we can see that on the one side, it is normal that many knee points from the entire Pareto fronts are not in their corresponding preference regions; on the other side, our aim of finding more fine-grained resolution in the preference region has been well achieved because only few solutions can be obtained in the preference region if we spread the population across the entire Pareto front. At the same time, one knee point from DI-1 on DTLZ2 is dominated by solutions from the corresponding AP-DI-1, which proves that AP-DI-MOEA can converge better than DI-MOEA because AP-DI-MOEA focuses on the preference region.

\begin{table}[!htbp]
\centering
\caption{Space and dominance relation of knee point from DI-MOEA and AP solutions on DTLZ problems.}
\begin{tabular}{|c|c|c|c|c|c|}
\hline
\multicolumn{2}{|c|}{Problem} & \multicolumn{2}{|c|}{DTLZ1}  & \multicolumn{2}{|c|}{DTLZ2}\\ 
\cline{1-6}
\multicolumn{2}{|c|}{\multirow{2}*{Algorithm}}& DI-1/ & DI-2/ & DI-1/ & DI-2/ \\
\multicolumn{2}{|c|}{ }& AP-DI-1 & AP-DI-2 & AP-DI-1 & AP-DI-2 \\
\hline
In & Incomparable & 29 & 27 & 10 & 13\\
\cline{2-6}
preference & Dominated & 0 & 0 & 1 & 0 \\
 \cline{2-6}
region & Dominating & 0 & 0 & 0 & 0 \\
\hline
Outside & Incomparable & 1 & 3 & 19 & 17 \\
\cline{2-6}
p-region & Dominated &  0 & 0 & 0 & 0 \\
\hline
\end{tabular}
\label{table-kneedtlz}
\end{table}

AP-DI-1 and AP-DI-2 have also been compared with NSGA-III in the same way. Table~\ref{table-kneedtlz_nsga3} shows the comparison result. For DTLZ1, the average number of solutions from NSGA-III in the corresponding preference regions from AP-DI-MOEA is six. Still, almost all knee solutions from NSGA-III are in the preference region. For DTLZ2, the average number of solutions from NSGA-III in the corresponding preference region from AP-DI-MOEA is less than one, while, in more than half of 30 runs, the knee points from NSGA-III are still in the preference region. To some extent, it can be concluded that the preference regions from AP-DI-MOEA are accurate. It can also be observed that AP-DI-1 behaves better than AP-DI-2 on DTLZ2, because two knee points from NSGA-III dominate the solutions from AP-DI-2.

\begin{table}[!htbp]
\centering
\caption{Space and dominance relation of knee point from NSGA-III and AP solutions on DTLZ problems.}
\begin{tabular}{|c|c|c|c|c|c|}
\hline
\multicolumn{2}{|c|}{Problem} & \multicolumn{2}{|c|}{DTLZ1}  & \multicolumn{2}{|c|}{DTLZ2}\\ 
\cline{1-6}
\multicolumn{2}{|c|}{\multirow{2}*{Algorithm}}& NSGA-III/ & NSGA-III/ & NSGA-III/ & NSGA-III/ \\
\multicolumn{2}{|c|}{ }& AP-DI-1 & AP-DI-2 & AP-DI-1 &AP-DI-2 \\
\hline
In & Incomparable & 30 & 29 & 14 & 17\\
\cline{2-6}
preference & Dominated & 0 & 0 & 1 & 1 \\
 \cline{2-6}
region & Dominating & 0 & 0 & 0 & 2 \\
\hline
Outside & Incomparable & 0 & 1 & 15 & 10 \\
\cline{2-6}
p-region & Dominated &  0 & 0 & 0 & 0 \\
\hline
\end{tabular}
\label{table-kneedtlz_nsga3}
\end{table} 

\begin{figure}[htbp]
\hspace{-0.58cm}
\subfigure[DTLZ1]{
    \begin{minipage}[t]{0.53\linewidth}
        \centering
        \includegraphics[width=2in]{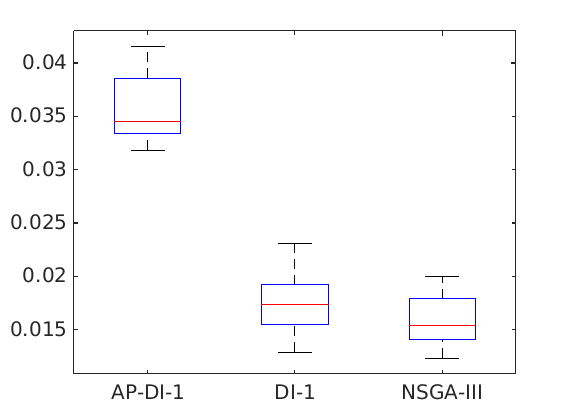}\\
        \vspace{0.02cm}
        \includegraphics[width=2in]{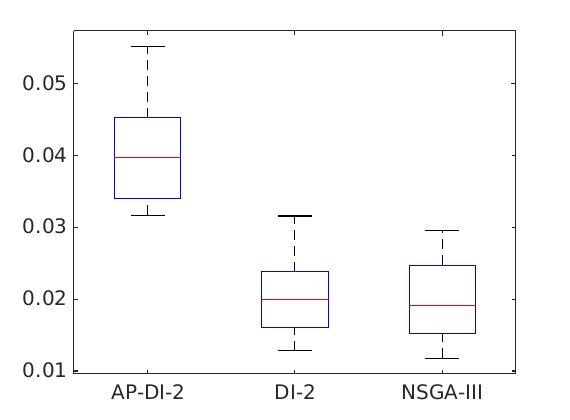}\\
        \vspace{0.02cm}
    \end{minipage}%
}%
\subfigure[DTLZ2]{
    \begin{minipage}[t]{0.53\linewidth}
        \centering
        \includegraphics[width=2in]{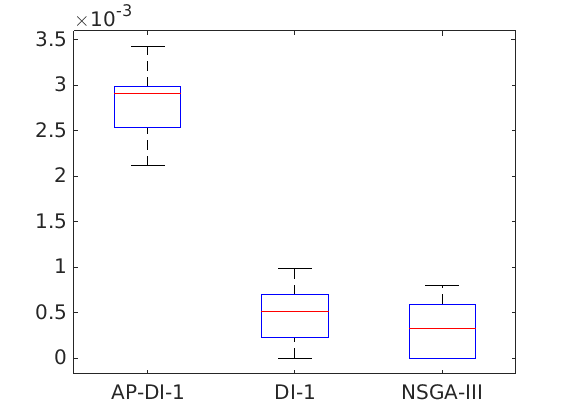}\\
        \vspace{0.02cm}
        \includegraphics[width=2in]{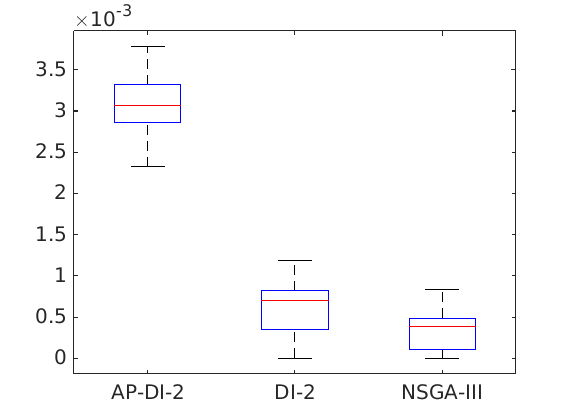}\\
        \vspace{0.02cm}
    \end{minipage}%
}%

\caption{Boxplots comparing the hypervolume values on DTLZ1 and DTLZ2.}
\label{box:dtlz}
\end{figure}

Similarly, we pick from DI-MOEA and NSGA-III solutions which are in the corresponding preference region of AP-DI-MOEA, and the hypervolume indicator value is compared between these solutions and AP solutions. It has been found that all hypervolume values of solutions from AP-DI-MOEA are better than those of solutions from DI-MOEA and NSGA-III. The left column of Figure~\ref{box:dtlz} shows box plots of the distribution of hypervolume values over 30 runs on DTLZ1, and the right column shows the hypervolume comparison on DTLZ2.

\begin{figure*}[htbp]
\hspace{-3.5cm}
\subfigure[DTLZ1 3 objective problem]{
    \begin{minipage}[t]{0.50\linewidth}
        \centering
        \includegraphics[height=2.6in,width=4.28in]{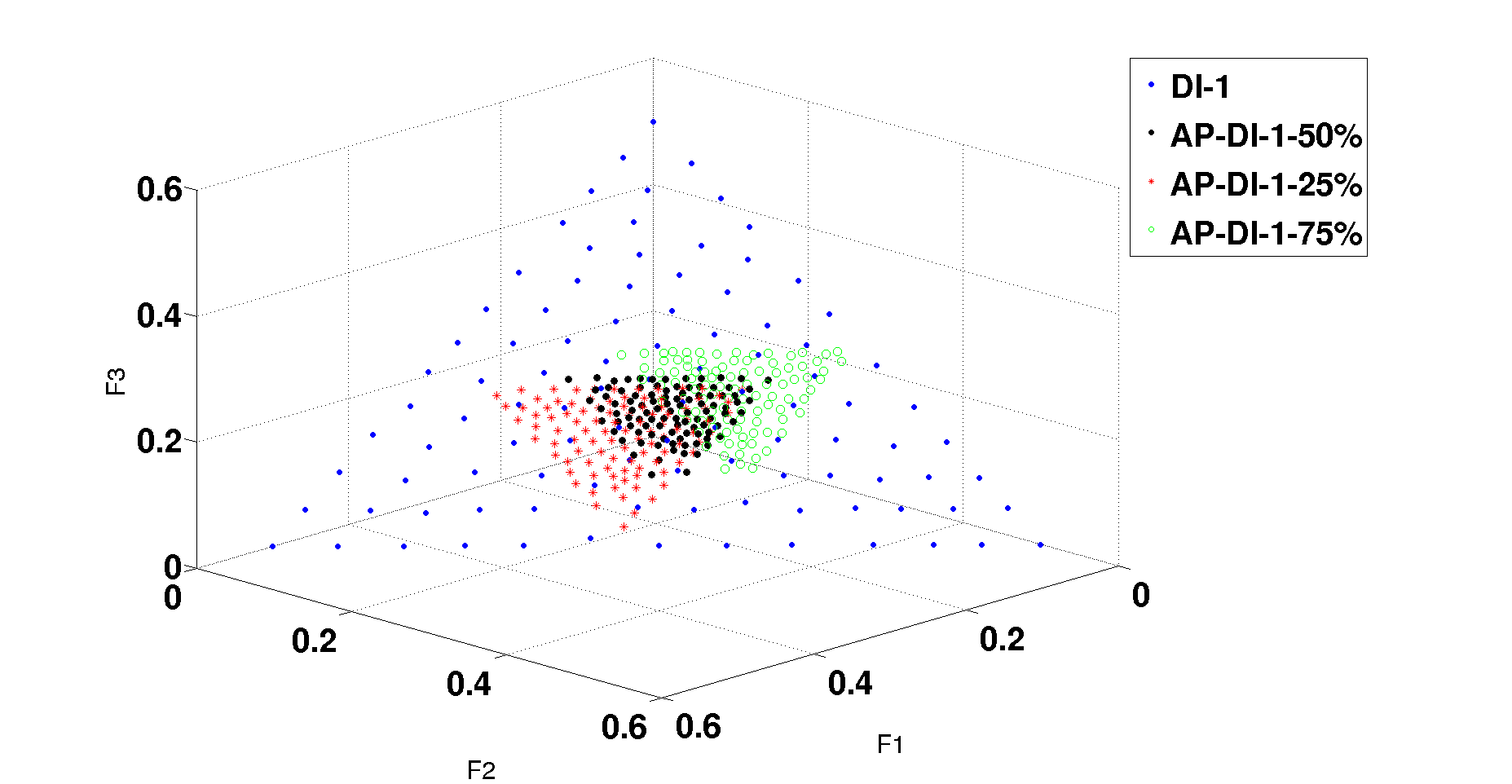}\\
        \includegraphics[height=2.6in,width=4.28in]{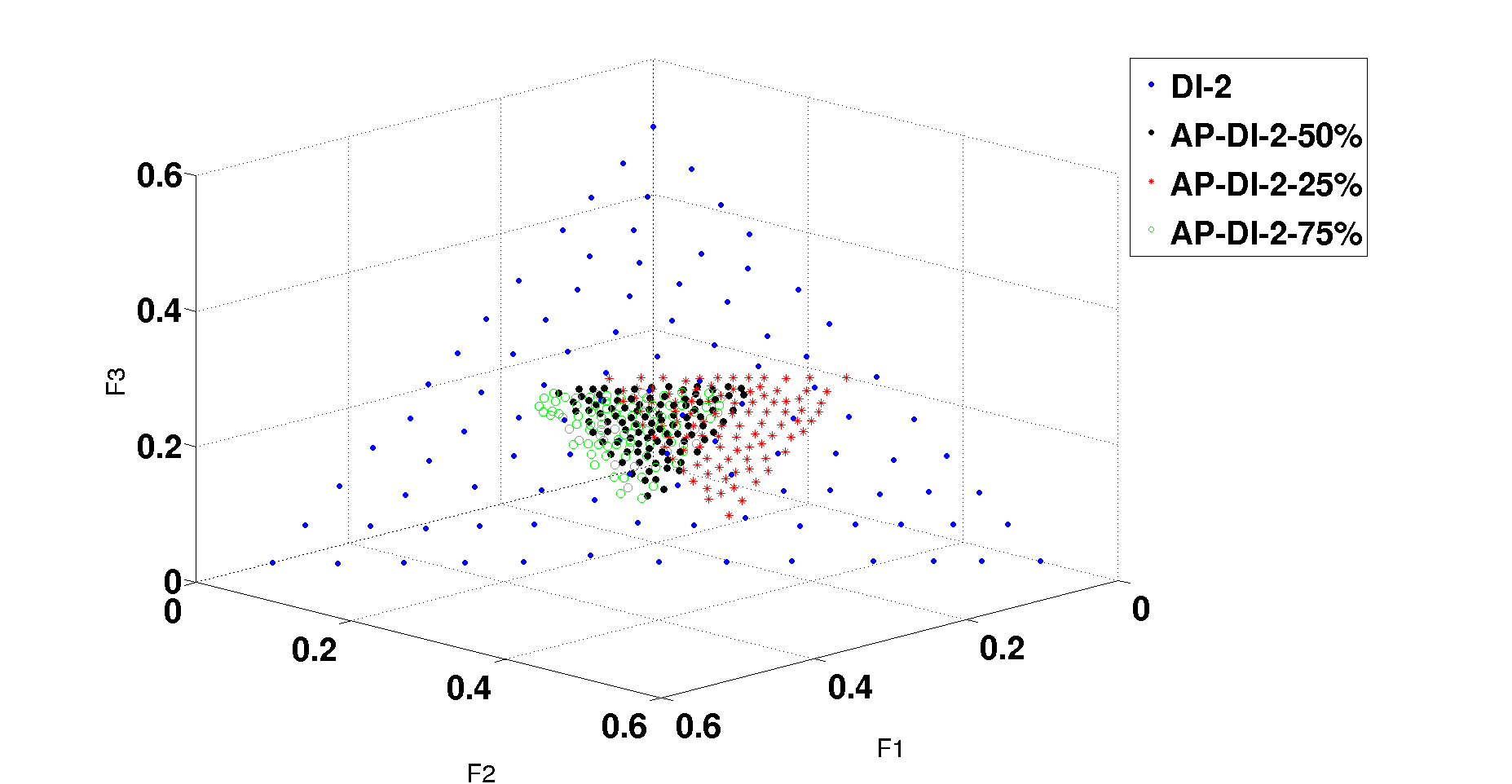}\\
    \end{minipage}%
}%
\hspace{1.5cm}
\subfigure[DTLZ2 3 objective problem]{
    \begin{minipage}[t]{0.50\linewidth}
        \centering
        \includegraphics[height=2.6in,width=4.2in]{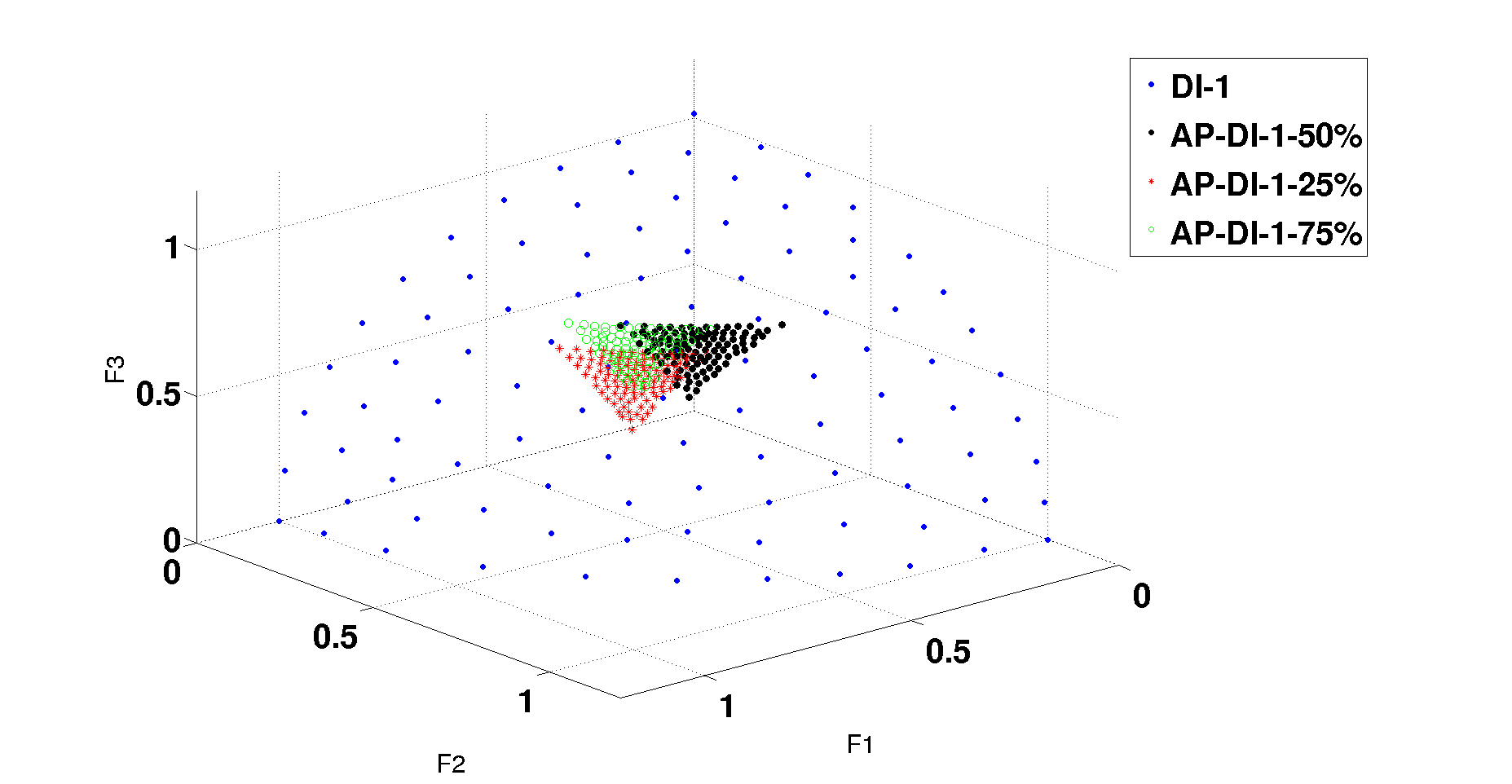}\\
        \includegraphics[height=2.6in,width=4.2in]{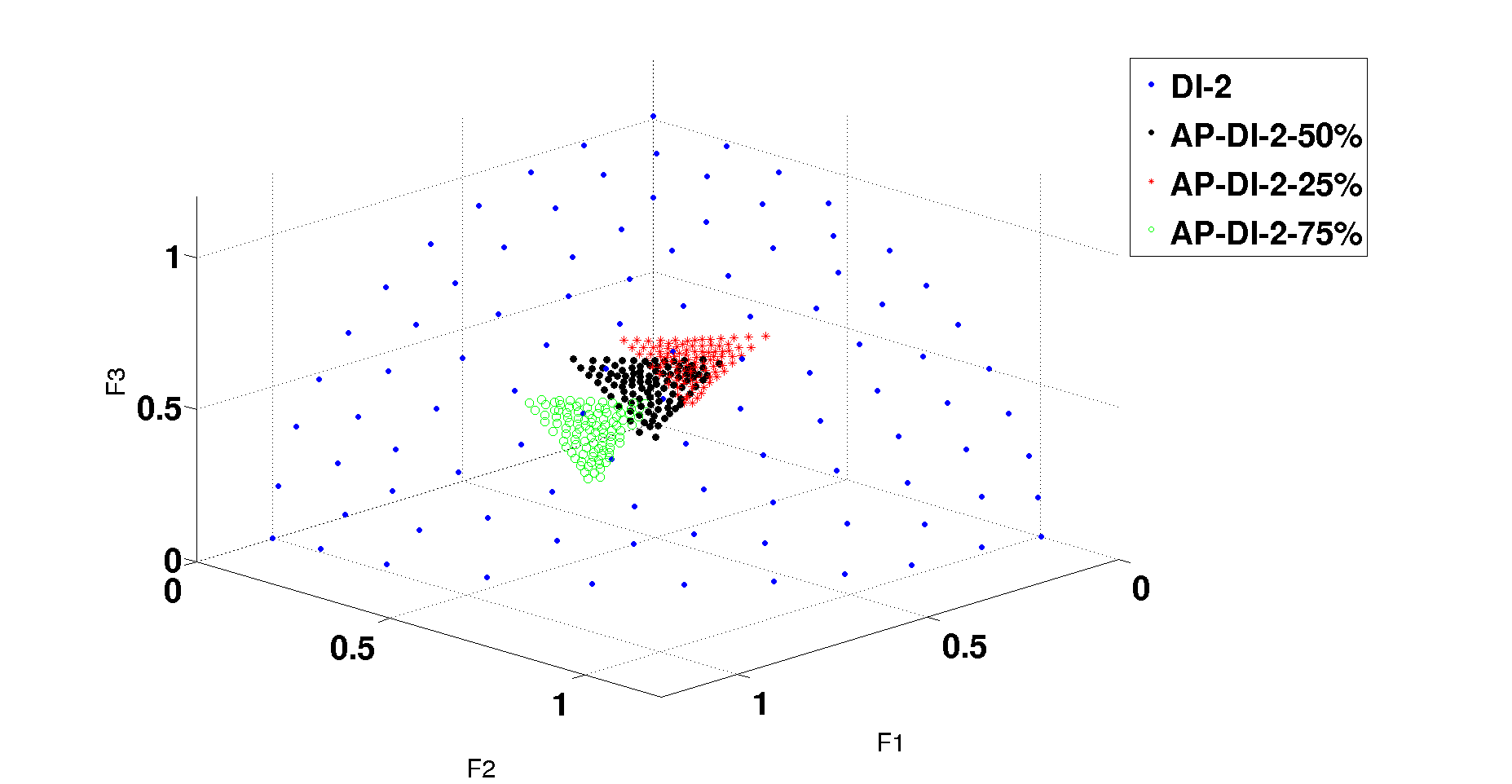}\\
    \end{minipage}%
}%
\caption{Pareto front approximation by different budgets generating initial Pareto front.}
\label{fig:dtlz-budget}
\end{figure*}

In our experiments, we decide half of the total budget is used to find an initial Pareto front because it turned out to be a good compromise: half budget for the initial Pareto front and another half budget for the solutions focusing on the preference region. We also run experiments using 25\% and 75\% of the total budget for the initial Pareto front. Figure~\ref{fig:dtlz-budget} presents the entire Pareto front from DI-MOEA and the Pareto front from AP-DI-MOEA with different budgets for the initial Pareto front. The left two images are on DTLZ1 and the right two images are on DTLZ2. The uppper two images are from DI-1 and AP-DI-1; the lower two images are from DI-2 and AP-DI-2. In the legend labels, 50\%, 25\% and 75\% indicate the budgets which are utilized to find the initial entire Pareto front. It can be observed that the preference region from AP-DI-MOEA with 50\% of budget are located on a better position than with 25\% and 75\% budgets, and the position of the preference region from AP-DI-MOEA with 50\% of budget is more stable. Therefore, in our algorithm, 50\% of budget is used before the generation of preference region.

\subsection{Experiments on Vehicle Fleet Maintenance Scheduling Optimization}
\begin{figure*}[htbp]
\hspace{-3.8cm}
\subfigure[DI-1 \& AP-DI-1]{
    \begin{minipage}[t]{0.5\linewidth}
        \centering
        \includegraphics[height=2.7in,width=4in]{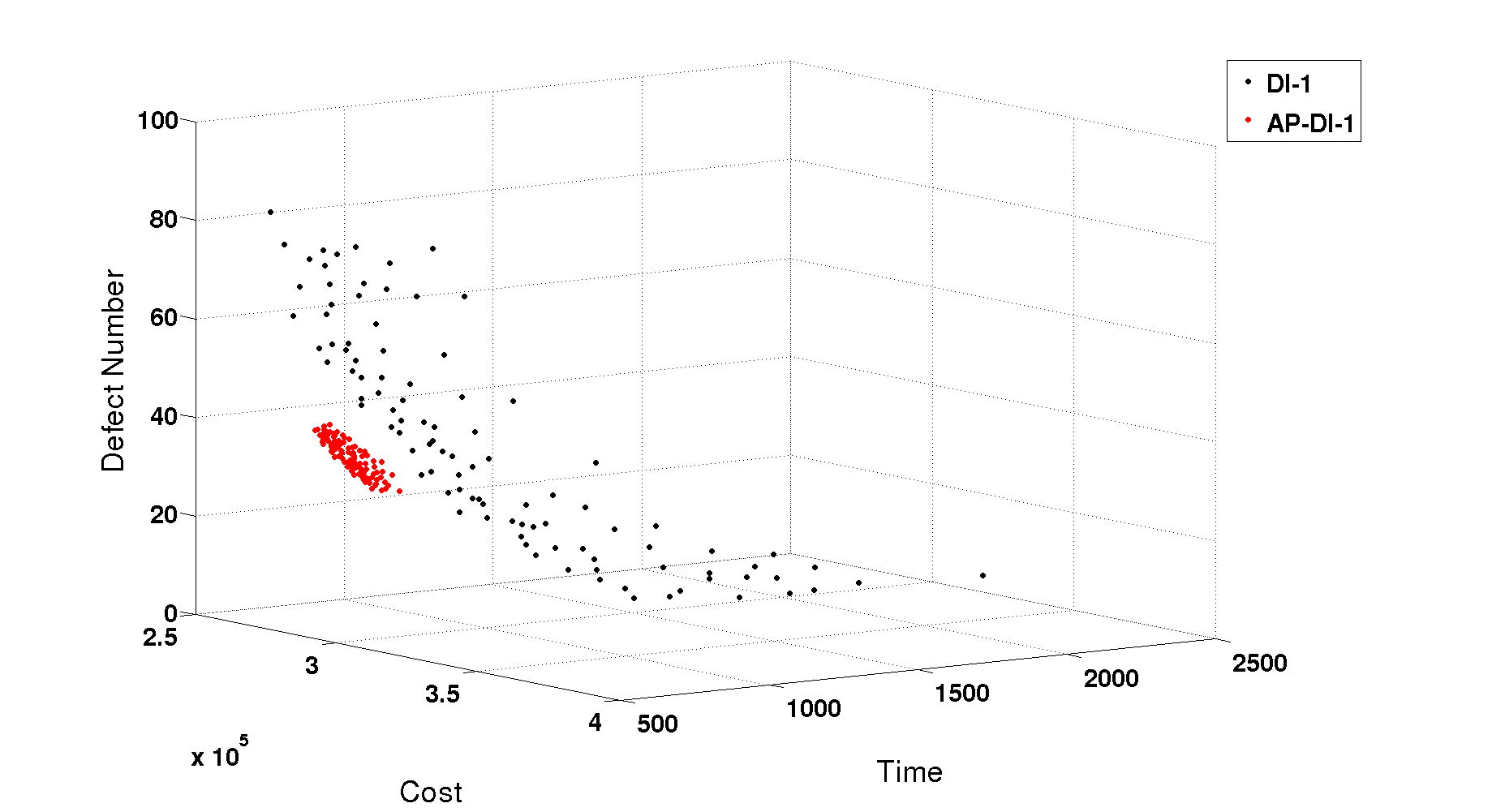}\\
    \end{minipage}%
}%
\hspace{3.2cm}
\subfigure[DI-2 \& AP-DI-2]{
    \begin{minipage}[t]{0.5\linewidth}
        \centering
        \includegraphics[height=2.7in,width=4in]{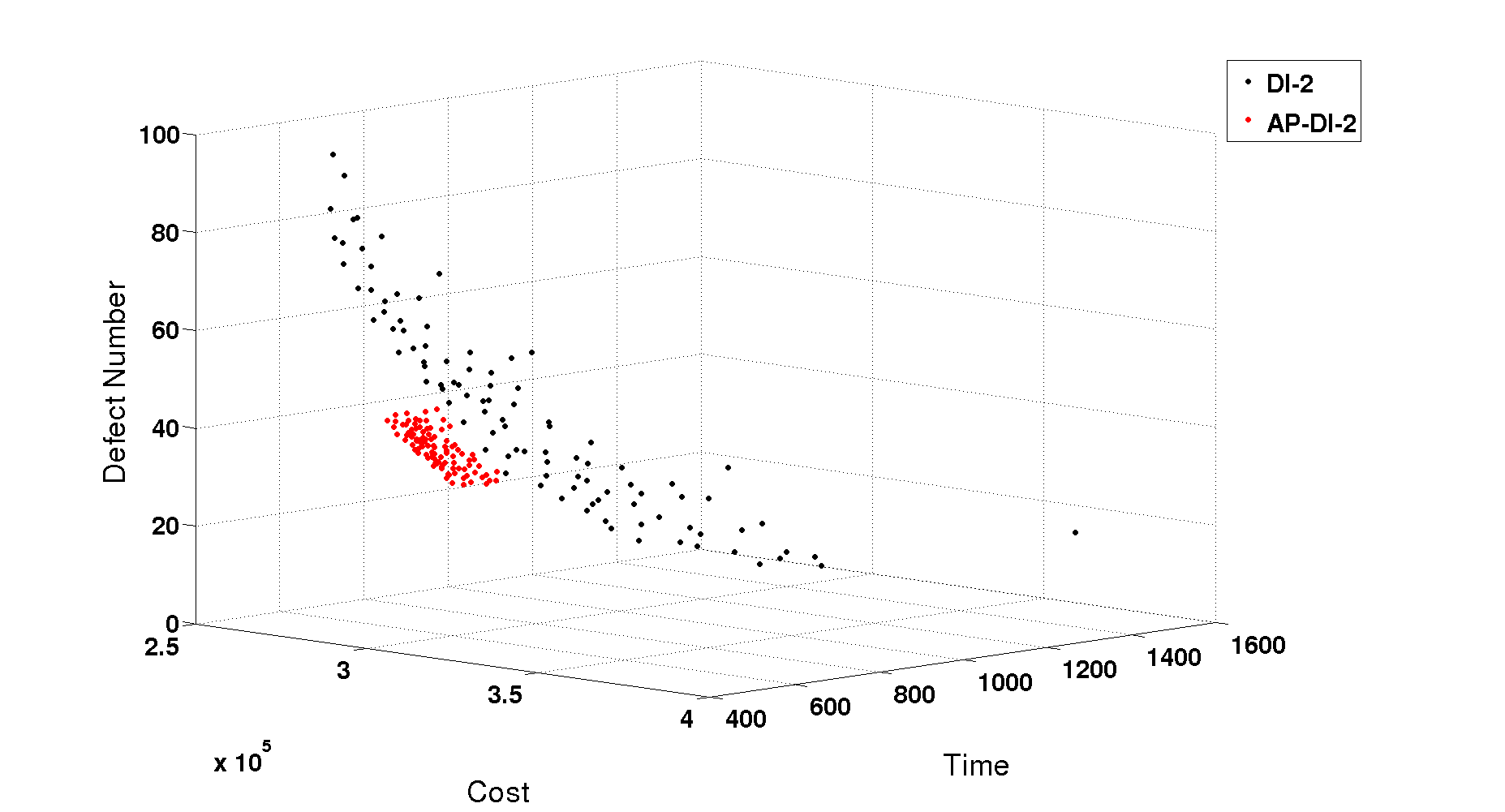}\\
    \end{minipage}%
}%
\caption{Pareto front approximation on VFMSO problem with 20 cars and 3 workshops.}
\label{fig:20cars}
\end{figure*}

\begin{figure*}[htbp]
\hspace{-3.8cm}
\subfigure[DI-1 \& AP-DI-1]{
    \begin{minipage}[t]{0.5\linewidth}
        \centering
        \includegraphics[height=2.7in,width=4in]{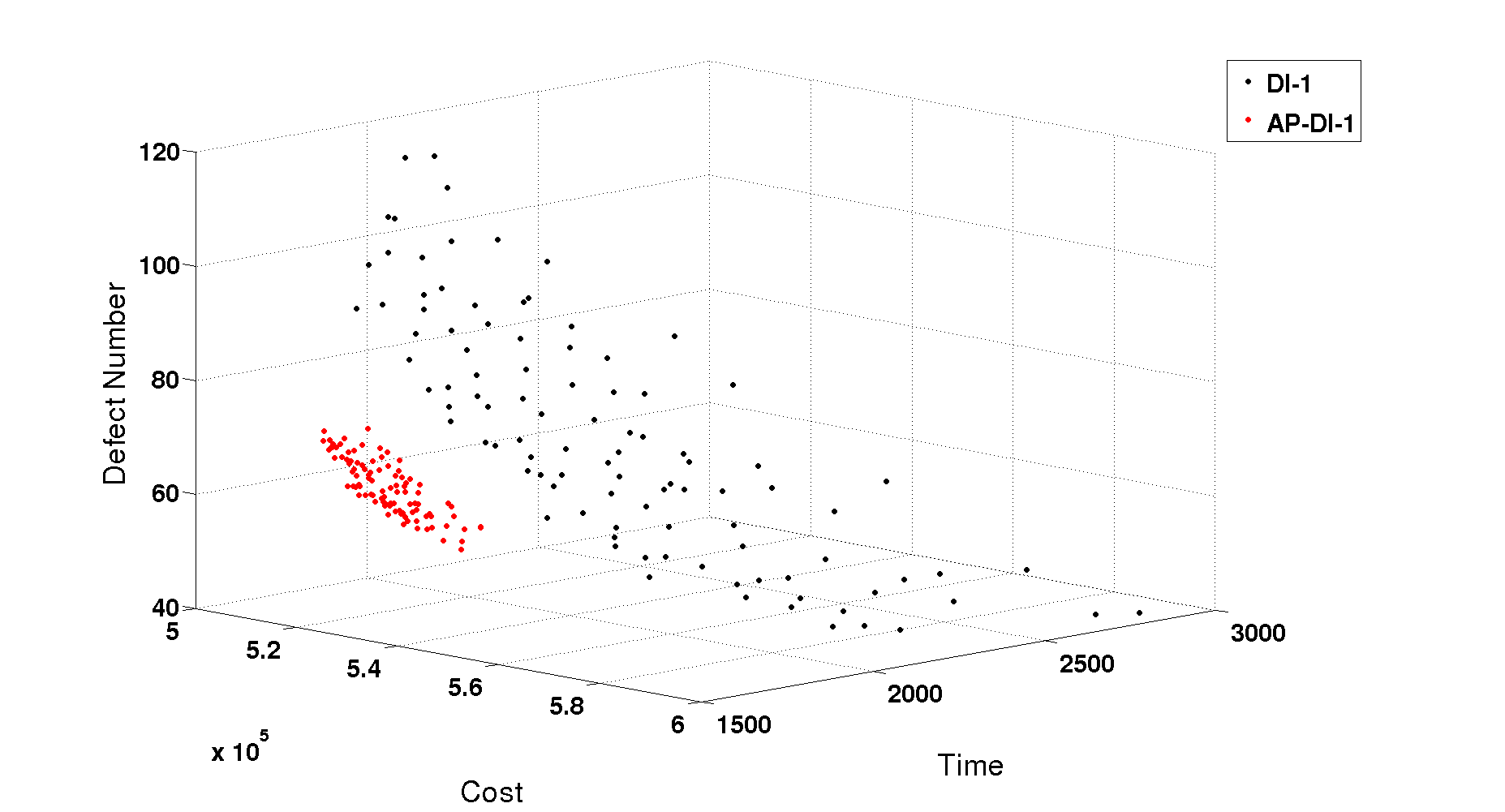}\\
    \end{minipage}%
}%
\hspace{3.2cm}
\subfigure[DI-2 \& AP-DI-2]{
    \begin{minipage}[t]{0.5\linewidth}
        \centering
        \includegraphics[height=2.7in,width=4in]{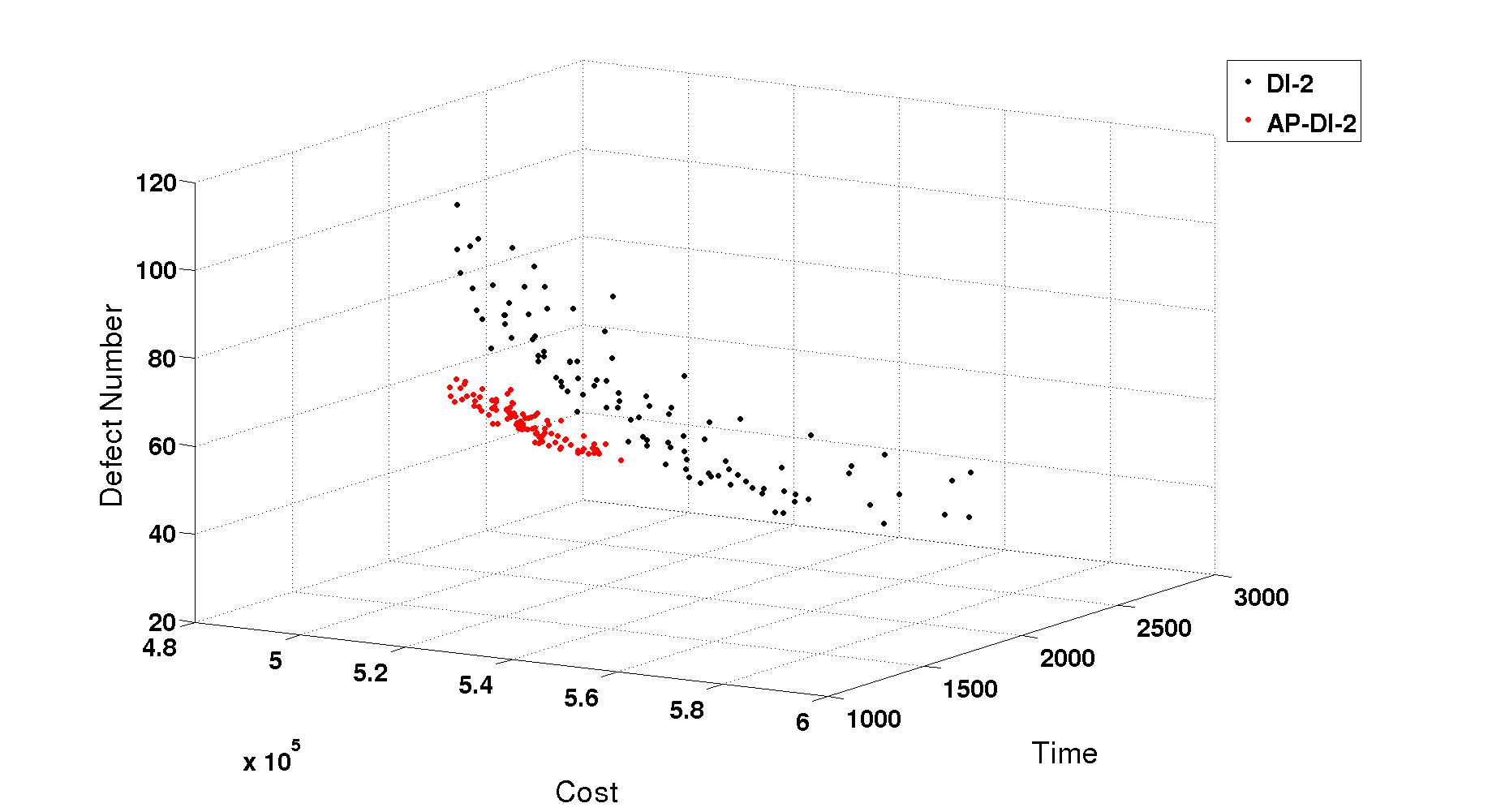}\\
    \end{minipage}%
}%
\caption{Pareto front approximation on VFMSO problem with 30 cars and 5 workshops.}
\label{fig:30cars}
\end{figure*}

The budget of $1200000$ evaluations has been used on the real-world application problems, and $600000$ of them are for the initial Pareto front. After that, the preference region is updated after every $50000$ evaluations.
The VFMSO problem has been tested with different sizes. Figure~\ref{fig:20cars} shows Pareto front approximations of a problem with $20$ cars and $3$ workshops (V1), and each car contains $13$ components: one engine, four springs, four brakes and four tires \cite{van2019modeling}. It can be observed that AP-DI-1 and AP-DI-2 can zoom in the entire Pareto front and find solutions in the preference region, at the same time, both AP-DI-1 and AP-DI-2 converge better than their corresponding DI-1 and DI-2. A similar conclusion can be drawn from Pareto fronts approximations of the problem with $30$ cars and $5$ workshops (V2) in Figure~\ref{fig:30cars}.

\begin{figure*}[htbp]
\hspace{-3.5cm}
\subfigure[V1]{
    \begin{minipage}[t]{0.5\linewidth}
        \centering
        \includegraphics[height=2.7in,width=4in]{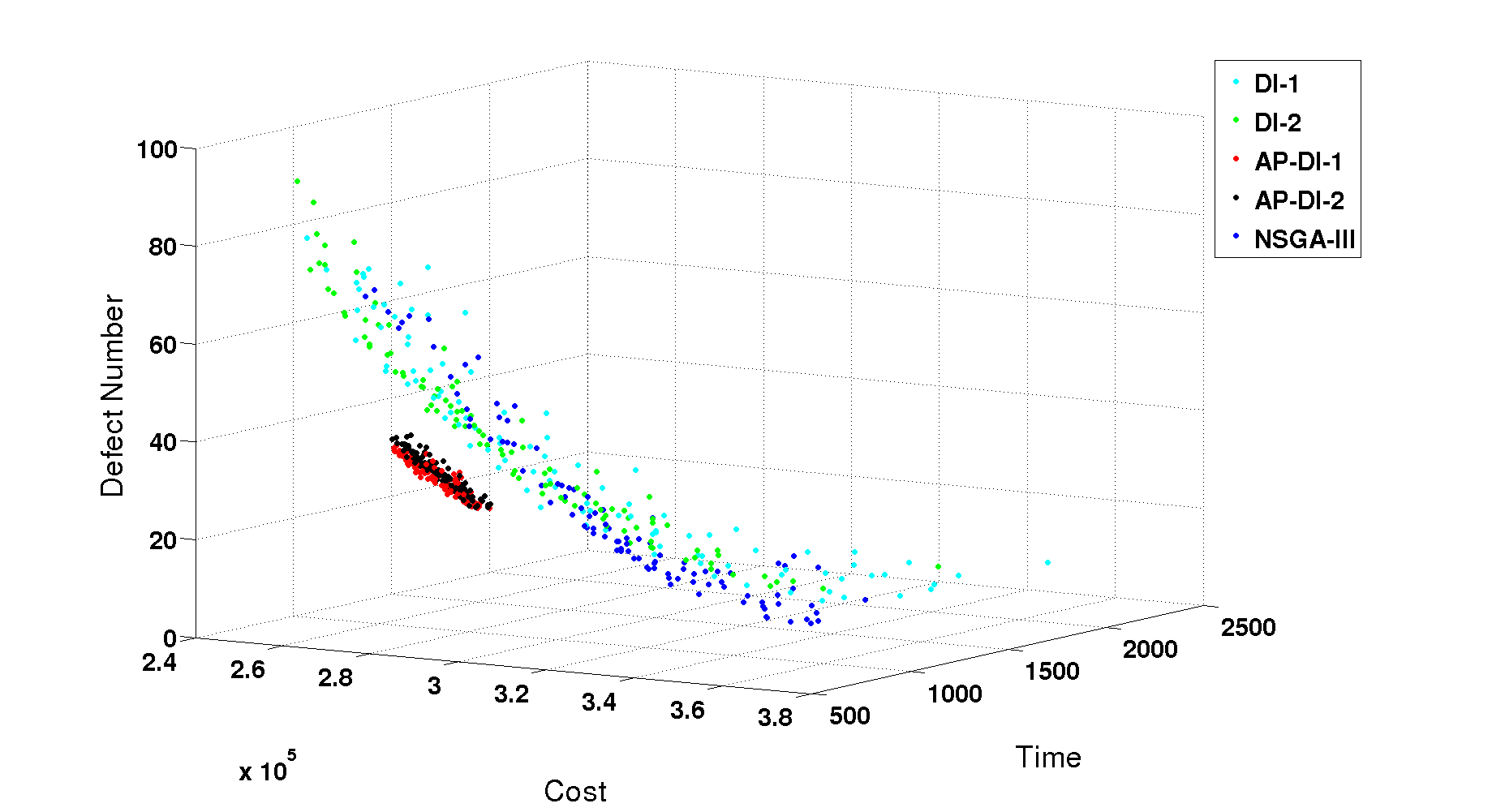}\\
    \end{minipage}%
}%
\hspace{3cm}
\subfigure[V2]{
    \begin{minipage}[t]{0.5\linewidth}
        \centering
        \includegraphics[height=2.7in,width=4in]{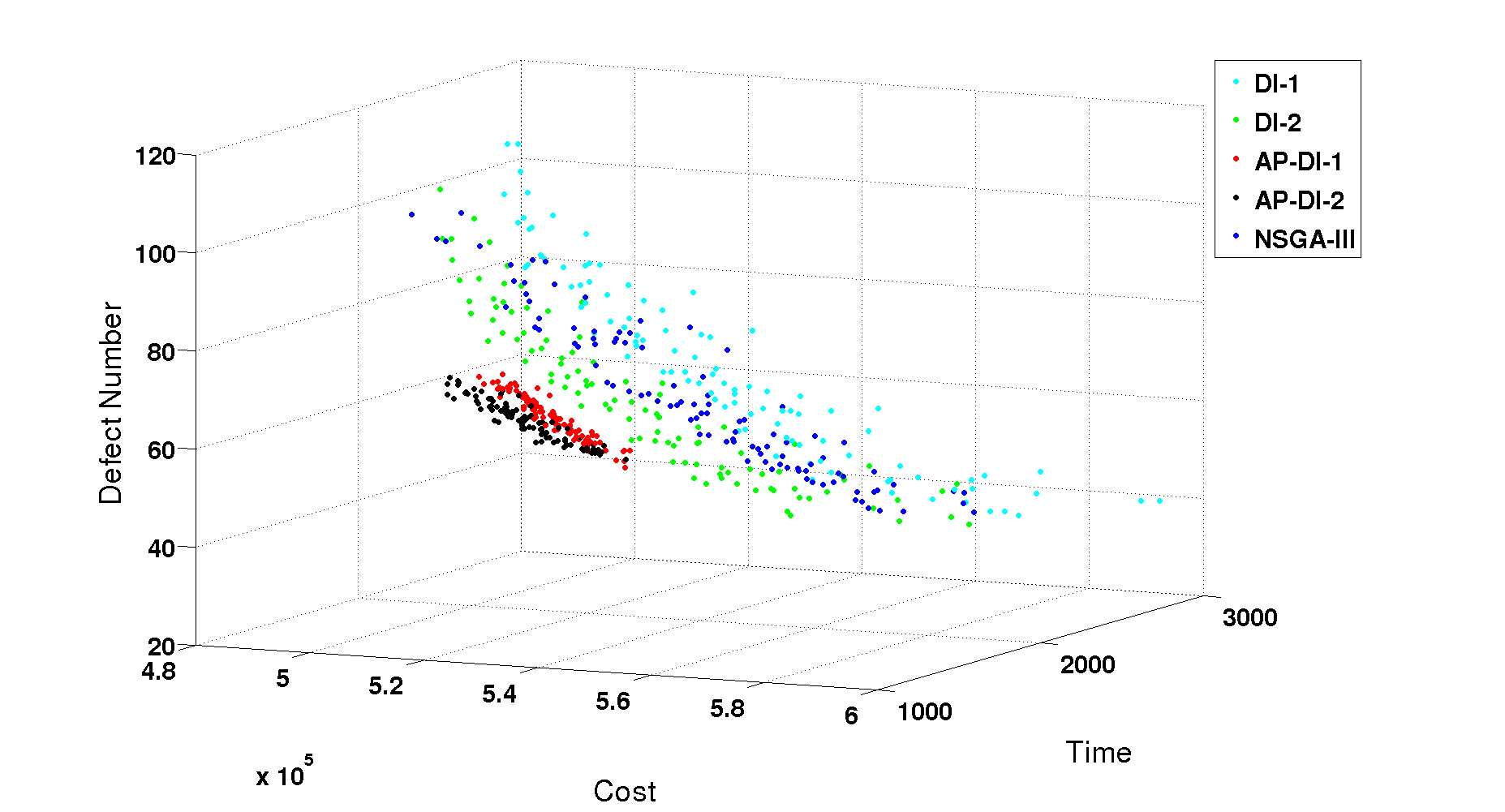}\\
    \end{minipage}%
}%

\caption{Pareto front approximation on VFMSO problem by DI-MOEA, AP-DI-MOEA and NSGA-III.}
\label{fig:2030cars}
\end{figure*}

In Figure~\ref{fig:2030cars}, We put the Pareto front approximations from DI-MOEA, AP-DI-MOEA and NSGA-III on V1 (left) and V2 (right) together. The behaviours of DI-1, DI-2 and NSGA-III are similar on V1, so are the behaviours of AP-DI-1 and AP-DI-2 on this problem. While, DI-2 and AP-DI-2 converge better than DI-1 and AP-DI-1 on V2 problem. The behaviour of NSGA-III is between that of DI-1 and DI-2.

Table~\ref{table-v1} gives the space and dominance relation of knee points from DI-MOEA and solutions from AP-DI-MOEA on these two VFMSO problems. For both problems, only few knee points from DI-MOEA are in the preference regions of AP-DI-MOEA, and the main reason is that the Pareto front of AP-DI-MOEA converges better than that of DI-MOEA, in some cases, the Pareto front of DI-MOEA cannot even reach the corresponding preference region. More importantly, it can be observed that most knee points from DI-MOEA, no matter whether in the preference region or outside of the preference region, are dominated by the solutions from AP-DI-MOEA. This phenomenon is even more obvious for the application problem with bigger size and run with the same budget as the smaller one: for V2, 90\% of knee points from DI-MOEA are dominated by the solutions from AP-DI-MOEA.

\begin{table}[!htbp]
\centering
\caption{Space and dominance relation of knee point from DI-MOEA and AP solutions on V1 and V2.}
\begin{tabular}{|c|c|c|c|c|c|}
\hline
\multicolumn{2}{|c|}{Problem} & \multicolumn{2}{|c|}{V1}  & \multicolumn{2}{|c|}{V2}\\ 
\cline{1-6}
\multicolumn{2}{|c|}{\multirow{2}*{Algorithm}}& DI-1/ & DI-2/ & DI-1/ & DI-2/ \\
\multicolumn{2}{|c|}{ }& AP-DI-1 & AP-DI-2 & AP-DI-1 & AP-DI-2 \\
\hline
In & Incomparable & 0 & 0 & 0 & 0\\
\cline{2-6}
preference & Dominated & 9 & 7 & 9 & 6 \\
 \cline{2-6}
region & Dominating & 0 & 0 & 0 &  0 \\
\hline
Outside & Incomparable &  4 & 9 & 3 & 3 \\
\cline{2-6}
p-region & Dominated & 17 & 14 & 18 & 21 \\
\hline
\end{tabular}
\label{table-v1}
\end{table}

Table~\ref{table-v1-nsga3} gives the space and dominance relation of knee points from NSGA-III and AP solutions. For both problems, again, most knee points from NSGA-III are not in the preference regions of AP-DI-MOEA. Some knee points from NSGA-III are dominated by AP solutions and most of them are incomparable with AP solutions.

\begin{table}[!htbp]
\centering
\caption{Space and dominance relation of knee point from NSGA-III and AP solutions on V1 and V2.}
\begin{tabular}{|c|c|c|c|c|c|}
\hline
\multicolumn{2}{|c|}{Problem} & \multicolumn{2}{|c|}{V1}  & \multicolumn{2}{|c|}{V2}\\ 
\cline{1-6}
\multicolumn{2}{|c|}{\multirow{2}*{Algorithm}}& NSGA-III/ & NSGA-III/ & NSGA-III/ & NSGA-III/ \\
\multicolumn{2}{|c|}{ }& AP-DI-1 & AP-DI-2 & AP-DI-1 & AP-DI-2 \\
\hline
In & Incomparable & 0 & 0 & 0 & 1\\
\cline{2-6}
preference & Dominated & 0 & 1 & 3& 2 \\
 \cline{2-6}
region & Dominating & 0 & 0 & 1 &  1 \\
\hline
Outside & Incomparable &  23 & 24 & 21 & 18 \\
\cline{2-6}
p-region & Dominated & 7 & 5 & 5 & 8 \\
\hline
\end{tabular}
\label{table-v1-nsga3}
\end{table} 

\section{CONCLUSIONS}
\label{sec:conclusion}
In this paper, a preference based multi-objective evolutionary algorithm, AP-DI-MOEA, is proposed. In the absence of explicitly provided preferences, the knee region is usually treated as the region of interest or preference region. Given this, AP-DI-MOEA can generate the knee region automatically and can find solutions with a more fine-grained resolution in the knee region. This has been demonstrated on the bi-objective problems ZDT1 and ZDT2, and the three objective problems DTLZ1 and DTLZ2. In the benchmark, the new approach was also proven to perform better than NSGA-III which was included in the benchmark as a state-of-the-art reference algorithm.

The research for the preference based algorithm was originally motivated by a real-world optimization problem, namely, Vehicle Fleet Maintenance Scheduling Optimization (VFMSO), which is described in this paper in a new formulation as a three objective discrete optimization problem. A customized set of operators (initialization, recombination, and mutation) is proposed for a multi-objective evolutionary algorithm with a selection strategy based on DI-MOEA and, respectively, AP-DI-MOEA. The experimental results of AP-DI-MOEA on two real-world application problem instances of different scales show that the newly proposed algorithm can generate preference regions automatically and it (in both cases) finds clearly better and more concentrated solution sets in the preference region than DI-MOEA. For completeness, it was also tested against NSGA-III and a better approximation in the preference region was observed by AP-DI-MOEA .

Since our real-world VFMSO problem is our core issue to be solved, and its Pareto front is convex, we did not consider problems with an irregular shape.
It would be an interesting question how to adapt the algorithm to problems with more irregular shapes. Besides, the proposed approach requires a definition of knee points. Future work will provide a more detailed comparison of different variants of methods to generate knee points, as they are briefly introduced in Section \ref{sec:literature}. In the application of maintenance scheduling, it will also be important to integrate robustness and uncertainty in the problem definition. It is desirable to generate schedules that are robust within a reasonable range of disruptions and uncertainties such as machine breakdowns and processing time variability. 

\section*{Acknowledgment}
This work is part of the research programme Smart Industry SI2016 with project name CIMPLO and project number 15465, which is (partly) financed by the Netherlands Organisation for Scientific Research (NWO).


\bibliography{elsarticle-template}

\end{document}